\documentclass{article} % For LaTeX2e
\usepackage{iclr2027_conference,times}

\usepackage{amsmath,amsfonts,bm}

\def\eqref#1{equation~\ref{#1}}
\def\1{\bm{1}}

\DeclareMathAlphabet{\mathsfit}{\encodingdefault}{\sfdefault}{m}{sl}
\SetMathAlphabet{\mathsfit}{bold}{\encodingdefault}{\sfdefault}{bx}{n}

\usepackage{hyperref}
\usepackage{url}
\usepackage{amssymb}          % \checkmark in the ablation-design table
\usepackage{amsmath}
\usepackage{amsthm}
\newtheorem{proposition}{Proposition}
\usepackage{booktabs}
\usepackage{graphicx}
\usepackage{wrapfig}
\usepackage{multirow}
\usepackage{xcolor}
\usepackage{colortbl}
\usepackage{algorithm}
\usepackage{algorithmic}
\definecolor{oursrow}{gray}{0.92}   % shading for the "ours" rows
\newcommand{\oc}{\cellcolor{oursrow}}   % shade one data cell

\title{Elite-Weighted Supervised Fine-tuning for Goal-Directed Molecular Optimization}

\author{Shiyun Wa$^{1,2}$ \quad Yifei Wang$^{1}$ \quad Anna G. Green$^{2}$ \quad
Simone Sciabola$^{1}$ \quad Ye Wang$^{1}$\thanks{Corresponding author:
\texttt{ye.wang@biogen.com}} \\[0.5em]
$^{1}$Research, Biogen, Cambridge, MA 02142, USA \\
$^{2}$Manning College of Information \& Computer Sciences, \\
University of Massachusetts Amherst, Amherst, MA 01003, USA
}

\newcommand{\best}[1]{\textbf{#1}}
\graphicspath{{../}{./}}

\iclrfinalcopy % Uncomment for camera-ready version, but NOT for submission.
\begin{document}

\maketitle
\lhead{Preprint. Under review.}

\begin{abstract}
Goal-directed optimization is essential for steering molecular generators to propose candidates with desired properties. However, it is often implemented with policy-gradient reinforcement learning, which requires a generation-trajectory log-probability whose form depends on the model architecture and generation procedure. This makes an optimizer difficult to reuse across architectures and conditional generative designs. Supervised fine-tuning needs none of that machinery, but its update is driven by a fixed dataset, so the reward never enters the update. We introduce Elite-Weighted Supervised Fine-tuning (EW-SFT), which uses reward to guide elite selection of high-scoring molecules, and updates the model by its own pretraining loss on that set. Ablations show that reward information is passed primarily through elite selection, rather than through continuous weighting within the selected set. Because the update consumes only scored molecules and the model's native loss, the same rule applies across autoregressive, masked-diffusion, and discrete-flow generators, and across de novo, motif-extension, and linker-design tasks.
Under a fixed budget of 3D shape alignment oracle calls on two kinase reference compounds, EW-SFT consistently outperforms the corresponding native optimizers. It further improves goal-directed optimization under a 2D similarity oracle on four held-out references and achieves comparable performance on a sample-efficiency benchmark without a trajectory-level RL formulation. These results demonstrate that EW-SFT is a unified and effective optimizer across molecular generators, design constraints, references, and oracles.
\end{abstract}

% ============================================================
\section{Introduction}

Generative models can propose novel small molecules directly with desired structures and properties rather than screening predefined libraries. Recent work has moved toward general-purpose molecular generators~\citep{safe-gpt,lee2025genmol,kaech2026refine} that support both de novo generation~\citep{GuacaMol}, where a complete molecule is built from scratch, and fragment-constrained generation~\citep{libinvent,drlinker,linkinvent,rush}, where a fixed substructure is retained and only the remainder is designed. However, optimizing such a generator toward a design objective is commonly done with policy-gradient reinforcement learning (RL)~\citep{10.1021/jacsau.4c00066}, which requires the log-probability assigned to the generation trajectory of each sampled molecule. This quantity depends on the architecture and generation procedure, so a policy-gradient objective must be specified for each architecture. For some two-pass generation procedures, the released implementation does not provide a joint policy over the complete generation trajectory. Meanwhile, keeping the policy-gradient update well behaved requires machinery whose settings have to be tuned~\citep{engstrom2020implementationmattersdeeppolicy,andrychowicz2020mattersonpolicyreinforcementlearning}.
In contrast, supervised fine-tuning updates the model through the same likelihood-based pretraining objective and needs none of this machinery. Its update is driven by given data samples, so the reward never reaches the model.

\begin{figure}[t]
    \centering
    \includegraphics[width=0.9\linewidth]{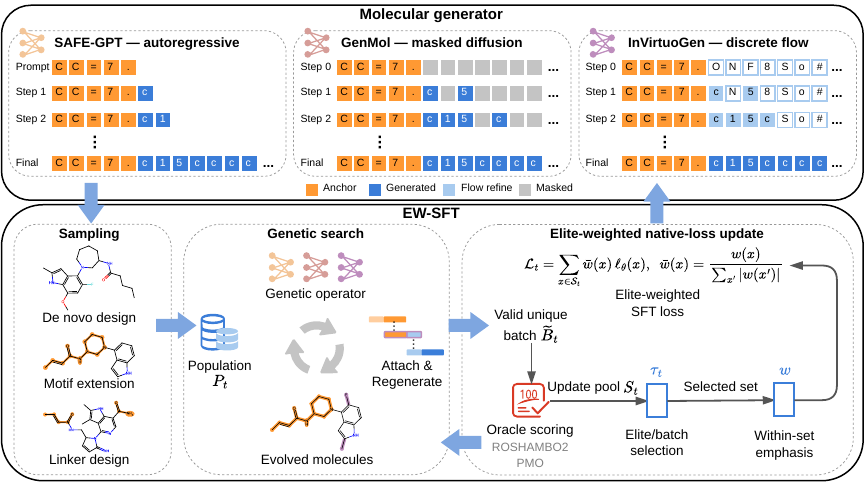}
    \caption{\textbf{The workflow of EW-SFT.} Three representative molecular generators first sample under task constraints to initialize a genetic population. Each model then acts as a genetic operator to regenerate new molecules, calling oracle to score and updating the population. Oracle scores determine which genetically evolved molecules are selected for the training set, and the chosen emphasis determines their weights in the native-loss update.}
    \label{fig:workflow}
\end{figure}

To fill this gap, we introduce Elite-Weighted Supervised Fine-tuning (EW-SFT), which lets the reward enter a supervised update through the composition of its training set. Each generator's genetic operator proposes molecules under its own unconditional or fragment-conditioned interface, a rolling \textit{elite} buffer retains the highest-scoring molecules discovered so far, and the model takes one weighted fine-tuning step on the selected set using its native pretraining loss (Figure~\ref{fig:workflow}). Prior molecular hill-climbing and GEGL methods also learn from selected high-scoring molecules~\citep{neil2018exploring,ahn2020guidingdeepmolecularoptimization}, but their original formulations target de novo optimization with exact-likelihood model refitting. EW-SFT instead extends this selection-based update across heterogeneous architectures and constrained generation procedures through each generator's native loss. We find that once elite selection has been applied, the \emph{emphasis} on each selected molecule can be reduced from a continuous advantage to a binary indicator. Because EW-SFT operates directly on completed, scored molecules, it requires neither a generation-trajectory log-probability nor PPO-style importance ratios, critics, or ratio clipping. It applies across autoregressive, masked-diffusion, and discrete-flow generators, and also enables optimization for SAFE-GPT two-pass linker generation without defining a joint policy over the two sampling passes.

We evaluate EW-SFT in ligand-based molecular design, where the objective is to generate molecules with good 3D similarity towards a known reference compound. This is based on the premise that molecules with similar 3D shape and pharmacophore arrangements can bind the same target and exhibit similar biological effects. Under a fixed budget of 3D similarity oracle calls on two kinase references, EW-SFT runs in all nine architecture--task combinations and consistently improves Top-1k similarity over the corresponding native optimizers. It further improves optimization under a 2D similarity oracle on the fragment-constrained generation benchmark~\citep{safe-gpt}, which recent generators have used only for frozen sampling. Finally, on the de novo Practical Molecular Optimization (PMO) benchmark~\citep{gao2022sample}, EW-SFT achieves performance comparable to a tuned policy-gradient method that carries more RL machinery. Our contributions are summarized as follows:

\begin{itemize}
  \item We propose EW-SFT, a general goal-directed update rule that applies across architectures and tasks using only scored molecules and each model's native pretraining loss.

   \item We show that elite selection, rather than continuous within-set weighting, is the effective channel through which reward reaches the update.

  \item We show that EW-SFT transfers to fragment-constrained design, where policy-gradient RL requires a task-specific formulation or a joint policy is not provided by the released generation procedure, and report the first goal-directed optimization results for generalist molecular generators under motif and linker constraints.

\end{itemize}

%%%%%%%%%
\section{Preliminaries}
\label{sec:prelim}

\paragraph{Molecular generation and optimization.}
Molecular generative models were long built one task at a time. Simplified Molecular Input Line Entry System (SMILES)-based generators~\citep{doi:10.1021/ci00057a005} are often trained for de novo design~\citep{Olivecrona2017,reinvent4}, and separate architectures were introduced for scaffold decoration~\citep{libinvent} and linker design~\citep{linkinvent,drlinker}. Sequential Attachment-based Fragment Embedding (SAFE)~\citep{safe-gpt} represents a molecule as a sequence of attachable fragments, enabling one pretrained generator to support multiple design tasks. Generalist generators built on this representation differ in architecture: SAFE-GPT~\citep{safe-gpt} is autoregressive, GenMol~\citep{lee2025genmol} is a masked diffusion model, and InVirtuoGen~\citep{kaech2026refine} is a discrete flow model. Their native pretraining objectives differ accordingly: SAFE-GPT uses an exact autoregressive negative log-likelihood, whereas GenMol and InVirtuoGen use likelihood-bound objectives. These models are evaluated by frozen sampling~\citep{safe-gpt} or by goal-directed optimization under an oracle-call budget~\citep{GuacaMol,gao2022sample}. Existing optimizers are paired with particular generator architectures and generation procedures and are not directly interchangeable across them.

\paragraph{Policy-gradient RL.}
Recent molecular optimizers commonly use trajectory-level policy gradients~\citep{reinvent4,10.1021/jacsau.4c00066,kaech2026refine,zhu2026graphgrpotraininggraphflow}, which scale the gradient of the log-probability assigned to a generation trajectory by its reward or advantage~\citep{Williams1992}. PPO-style variants additionally use an importance ratio between the current and behavior policies to reuse sampled trajectories~\citep{ppo}. 
For an autoregressive generator, the generation trajectory factorizes over token decisions. For iterative generators~\cite{kaech2026refine,zhu2026graphgrpotraininggraphflow}, the trajectory and its probability depend on the architecture. Generation procedure can add an additional dependence: SAFE-GPT linker design samples the two anchor-constrained sequences in separate passes and joins them, so policy-gradient RL requires a joint policy over the complete two-pass generation trajectory.
Even when a policy-gradient objective can be formed, keeping its update well behaved depends heavily on auxiliary design choices. Code-level choices such as value-function clipping and reward scaling can affect final reward more than the training algorithm choice, and their best settings are environment-dependent~\citep{engstrom2020implementationmattersdeeppolicy,andrychowicz2020mattersonpolicyreinforcementlearning}. In molecular optimization, systematic studies of REINFORCE-based methods found that common prior-likelihood regularization need not improve optimization \citep{10.1021/acs.jcim.5c02053}. Prior anchors, KL penalties, diversity filters, and replay can improve particular settings but introduce many hyperparameters. Their optima shift across objectives, tuning consumes a finite oracle budget~\citep{gao2022sample}, and the resulting updates can still collapse or require extensive tuning~\citep{Thomas2022,10.1021/jacsau.4c00066}.

\paragraph{Supervised fine-tuning.}
Supervised fine-tuning instead updates a generator on completed molecules under its own native pretraining loss. It therefore does not require a trajectory log-probability or a policy-gradient formulation~\citep{segler2017generatingfocussedmoleculelibraries,gupta2018}. But for a given dataset, the update has no oracle feedback. A higher-scoring molecule and a lower-scoring one contribute identically, and where the data comes from is a decision made outside the update~\citep{neil2018exploring}.

\paragraph{Reward-guided supervised fine-tuning.}
These observations motivate a reward-guided native-loss update. After candidate molecules have been proposed and scored, the shared fine-tuning update should depend only on completed molecules, their oracle scores, and each generator's native pretraining loss, rather than on the generation trajectory that produced them. A weighted native-loss update provides such an interface. Three design decisions then remain --- where the training molecules come from, which of them are selected, and how strongly each selected molecule counts. Our ablations test these three roles separately and show that genetic search supplies proposals beyond unguided sampling, elite selection carries the useful reward signal, and continuous within-buffer weighting is unnecessary.

\section{Method}
\label{sec:method}

EW-SFT instantiates the weighted native-loss update identified in Section~\ref{sec:prelim}. It adapts to each generator through its native genetic operator $G$ and pretraining loss $\ell_\theta$ while applying the same reward-guided selection and weighting formulation. Algorithm~\ref{alg:ewsft} gives the GenMol instantiation with elite selection, the other two backbones differ only in $G$ and $\ell_\theta$ (Appendix~\ref{app:impl}).

\subsection{Problem setting}
Given a pretrained generative model $p_\theta$ conditioned on structural constraint $C$ and a budget of $N$ oracle calls to $f$, our goal is to generate the highest scoring compounds under $f$ within this evaluation budget. Following the taxonomy of \cite{safe-gpt}, we use three generative design tasks: de novo design ($C = \varnothing$), motif extension ($C$ is one fragment that must be retained), and linker design ($C$ is two fragments that must both be retained and joined). We use $\mathcal{M}_C$ to denote the feasible set of molecules that qualify under $C$. Frozen sampling, which draws $N$ samples from the frozen $p_\theta$ and scores them with $\theta$ unchanged, is what current generators support natively under constraints.

\subsection{Elite-Weighted Supervised Fine-tuning}
\label{sec:ew-sft}

We apply EW-SFT to three pretrained molecular generators with distinct generation mechanisms and likelihood-based native pretraining losses $\ell_\theta$: SAFE-GPT \citep{safe-gpt} (autoregressive, next-token cross-entropy), GenMol \citep{lee2025genmol} (masked diffusion, Negative Evidence Lower Bound (NELBO) loss~\citep{sahoo2024simpleeffectivemaskeddiffusion}), and InVirtuoGen \citep{kaech2026refine} (discrete flow, time-weighted flow-matching NELBO). EW-SFT replaces their corresponding native optimizers---PPO~\citep{ppo}, a genetic algorithm (GA)~\citep{lee2025genmol}, and Genetic-PPO, respectively--- with its reward-guided weighted native-loss update.

\begin{algorithm}[ht]
\caption{EW-SFT (GenMol).}
\label{alg:ewsft}
\begin{algorithmic}
\REQUIRE pretrained $p_\theta$; constraint $C$; oracle $f$; budget $N$; buffer size $K$; population size $|P|$
\STATE $\tilde{B}_0 \sim p_\theta(\cdot \mid C)$;\quad $f(\tilde{B}_0)$;\quad $P_0 \gets \mathrm{Cut}(\tilde{B}_0)$;\quad $E_0 \gets \mathrm{Top}\text{-}K(\tilde{B}_0)$;\quad $n \gets |\tilde{B}_0|$
\COMMENT{Cold start}
\FOR{$t = 1, 2, \dots$ \textbf{while} $n < N$}
  \STATE $\tilde{B}_t \gets G(P_{t-1})$
   \COMMENT{Attach fragment pairs, re-mask one slot, keep valid and unique}
  \STATE $f(\tilde{B}_t)$;\quad $n \gets n + |\tilde{B}_t|$
  \STATE $P_t \gets \mathrm{Top}\text{-}|P|\big(P_{t-1} \cup \mathrm{Cut}(\tilde{B}_t)\big)$;\quad
   $\mathcal{S}_t \gets \tilde{B}_t \cup E_{t-1}$;\quad $E_t \gets \mathrm{Top}\text{-}K(\mathcal{S}_t)$ \COMMENT{Update}
  \IF{$E_t \neq E_{t-1}$}
    \STATE $\theta_t \gets \theta_{t-1} - \eta \nabla_\theta \sum_{x \in \mathcal{S}_t} \bar{w}(x)\, \ell_\theta(x)$
     \COMMENT{Eq.~\ref{eq:sft} and ~\ref{eq:weight-factor}}
  \ENDIF
\ENDFOR
\RETURN $\bigcup_{t \geq 0} \tilde{B}_t$
\COMMENT{Scored molecules}
\end{algorithmic}
\end{algorithm}

\paragraph{Genetic search.} Each pretrained generator acts as an architecture-specific genetic operator $G$. It first samples a batch $B_0 \subset \mathcal{M}_C$ to initialize the population $P_0$. At each subsequent round, $G$ regenerates the batch $B_t$ from the population $P_{t-1}$. We adapt each model's native de novo genetic mechanism to the fragment-constrained tasks. The population stores high-scoring fragments or molecules, and constrained tokens are never mutated. The implementation details are in Appendix~\ref{app:impl}. Only the valid and unique molecules in batch $\tilde{B}_t \subset B_t$ are scored, consuming $|\tilde{B}_t|$ of the budget. Molecules not valid for the oracle calculation still consume a call but are excluded from training.

\paragraph{Elite-weighted native-loss update.} At round $t$, let $\mathcal{S}_t$ denote the scored update pool. EW-SFT updates the generator by one gradient step of weighted native-loss fine-tuning over this pool,
\begin{equation}
\theta_{t+1} = \theta_t - \eta \nabla_\theta
  \sum_{x \in \mathcal{S}_t} \bar{w}(x)\, \ell_\theta(x),
\qquad
\bar{w}(x) = \frac{w(x)}{\sum_{x' \in \mathcal{S}_t} |w(x')|},
\label{eq:sft}
\end{equation}
where $\ell_\theta$ is the model's own pretraining loss and $\eta$ is the learning rate. Because every $x\in\mathcal{S}_t$ has already met the structural constraint, the update evaluates each model's native pretraining loss on the completed molecule $x$ itself. The oracle reaches the update through the weight $w$, which we write as an elite indicator gating a binary or continuous emphasis:
\begin{equation}
w(x) \;=\; \mathbf{1}\!\left[\,f(x) \geq \tau_t\,\right] \cdot
\begin{cases}
1 & \text{binary,}\\[2pt]
\hat{A}(x) & \text{signed,}\\[2pt]
\max\!\big(\hat{A}(x),\, 0\big) & \text{positive,}
\end{cases}
\qquad
\hat{A}(x) = \frac{f(x) - \mu_t}{\sigma_t + \epsilon},
\label{eq:weight-factor}
\end{equation}

Here, $\mu_t$ and $\sigma_t$ are the mean and standard deviation of $f$ over the selected molecules, and $\epsilon$ is a small constant. The selection scheme determines both the update pool $\mathcal{S}_t$ and the selected subset within it. Under batch selection, $\mathcal{S}_t = \tilde{B}_t$ and $\tau_t = -\infty$, so all valid molecules in the current round are selected. Under elite selection, EW-SFT maintains a rolling \emph{elite buffer} $E_t$ containing the $K$ highest-scoring molecules seen through round $t$, initialized as $E_0 = \mathrm{Top}\text{-}K(\tilde{B}_0)$. It sets $\mathcal{S}_t = \tilde{B}_t \cup E_{t-1}$ and $E_t = \mathrm{Top}\text{-}K(\mathcal{S}_t)$. The corresponding selection threshold is $\tau_t = \min_{x \in E_t} f(x)$, which selects the elite buffer. $\tau_t$ rises monotonically as better molecules are found, and each sample competes against the run's whole history rather than against its own round.

Because the three generators maintain genetic populations at different granularity, we specify $E_t$ by function rather than by implementation, which is a set of $K$ whole molecules, ranked by their own oracle score and refreshed across rounds. GenMol and SAFE-GPT evolve populations of fragments, scored by the mean over the molecules a fragment was cut from, so no complete winner survives a round and $E_t$ must be maintained separately. InVirtuoGen's genetic population instead holds whole molecules ranked by their own $f$, which already satisfies this definition, and therefore we treat it as that model's elite buffer without adding the second one (Table~\ref{tab:ablation-design}).

Equations~\ref{eq:sft} and~\ref{eq:weight-factor} define different update rules through the choice of selection threshold and weighting scheme, which we compare in Section~\ref{sec:ablation}. We use the elite threshold with binary weighting, denoted elite$+$binary. In this setting, $\bar{w}(x)=1/|E_t|$ for molecules in $E_t$ and zero otherwise, so Equation~\ref{eq:sft} reduces exactly to the mean pretraining loss over $E_t$, corresponding to one step of standard fine-tuning on the best $K$ molecules found so far. This simple rule is sufficient because the oracle scores determine which molecules are retained in $E_t$, and the gradient update only learns from those selected molecules. Positive or signed weighting instead assigns different importance to molecules within the elite set. A signed emphasis potentially makes the objective unbounded below, since a negative weight on an unbounded-above loss may let gradient descent reduce the loss by pushing away disfavoured molecules rather than toward good ones (Proposition~\ref{prop:unbounded} and Figure~\ref{fig:signed-divergence}). Since the model is updated only when $E_t \neq E_{t-1}$, a stalled search does not repeatedly fit the same buffer, eliminating the need for an explicit early-stopping rule (Appendix~\ref{app:impl}).

% \begin{wrapfigure}[17]{r}{0.46\textwidth}
% \vspace{-\intextsep}
% \centering
% \includegraphics[width=0.43\textwidth]{figures/signed_divergence.pdf}
% \caption{\textbf{Signed emphasis drives the SFT loss below zero.} The signed per-round training loss crosses zero and keeps descending, every other setting stays bounded below.}
% \label{fig:signed-divergence}
% \end{wrapfigure}
% \paragraph{Why the emphasis is not signed.} A variant sets the emphasis to a batch-normalized advantage $\hat{A}$, which is negative for below-average samples. It is not merely worse empirically, but also unbounded. Gradient descent can therefore reduce the unbounded-below loss by pushing away disfavoured samples rather than toward the good ones. Figure~\ref{fig:signed-divergence} shows the failure mode. This update rule's property applies to all three models.

% \begin{proposition}[Signed emphasis makes the objective unbounded below]
% \label{prop:unbounded}
% Fix a batch $\tilde{B}_t$ under the signed emphasis, so that $w(x) = \hat{A}(x)$ and some $x^- \in \tilde{B}_t$ has $w(x^-) < 0$. Since
% $\ell_\theta(x^-) = -\log p_\theta(x^-) \in [0, \infty)$ is unbounded above, driving $p_\theta(x^-) \to 0$ while holding the remaining terms fixed lets $w(x^-)\,\ell_\theta(x^-) \to -\infty$ with every other term finite, so the weighted objective $\sum_{x \in \tilde{B}_t} w(x)\, \ell_\theta(x)$ is unbounded below. Restricting to $w \geq 0$ makes every term a non-negative multiple of a non-negative loss, so the objective is bounded below by $0$.
% \end{proposition}

\section{Experiments}

\subsection{3D Shape Similarity-Directed Optimization}
\label{sec:3d}

\paragraph{Setup.} We optimize the 3D shape similarity of generated molecules to a known reference compound under each structural constraint, using ROSHAMBO2~\citep{roshambo2}, an open-source 3D similarity tool, as the oracle with a budget of 20k calls. We use BMS-986195, a Bruton's tyrosine kinase (BTK) inhibitor~\citep{watterson2019discovery}, and Zasocitinib (TAK-279), a tyrosine kinase 2 (TYK2) inhibitor~\citep{10.1021/acs.jmedchem.3c00600}, as reference compounds. Table~\ref{tab:anchors} displays the fragment anchors for both. Within each backbone, EW-SFT replaces the native optimizer with all other settings held fixed. Because the goal is to find a high-scoring, novel and diverse pool of molecules, our primary metric is the mean oracle value of the best $1$k molecules. Every cell is measured against frozen sampling, drawing the same 20k molecules over 3 runs from the frozen model and scoring them unchanged, which is what current generators support natively under fragment constraints. Because no prior work has optimized these generalist generators under such constraints, we additionally adapt their own de novo optimizers to the constrained interface for motif and linker tasks. REINVENT-family models\footnote{REINVENT~\citep{Olivecrona2017,reinvent4} for de novo design, LibINVENT~\citep{libinvent} for motif extension, and Link-INVENT~\citep{linkinvent} for linker design} are per-task references only, since their convergence needs substantially more oracle calls~\citep{wa2026advancingligandbasedvirtualscreening}. Frozen sample quality is characterized in Appendix~\ref{app:frozen}, and detailed oracle settings and hyperparameters in Appendix~\ref{app:impl}.

\begin{figure}[ht]
\centering
\includegraphics[width=0.8\textwidth]{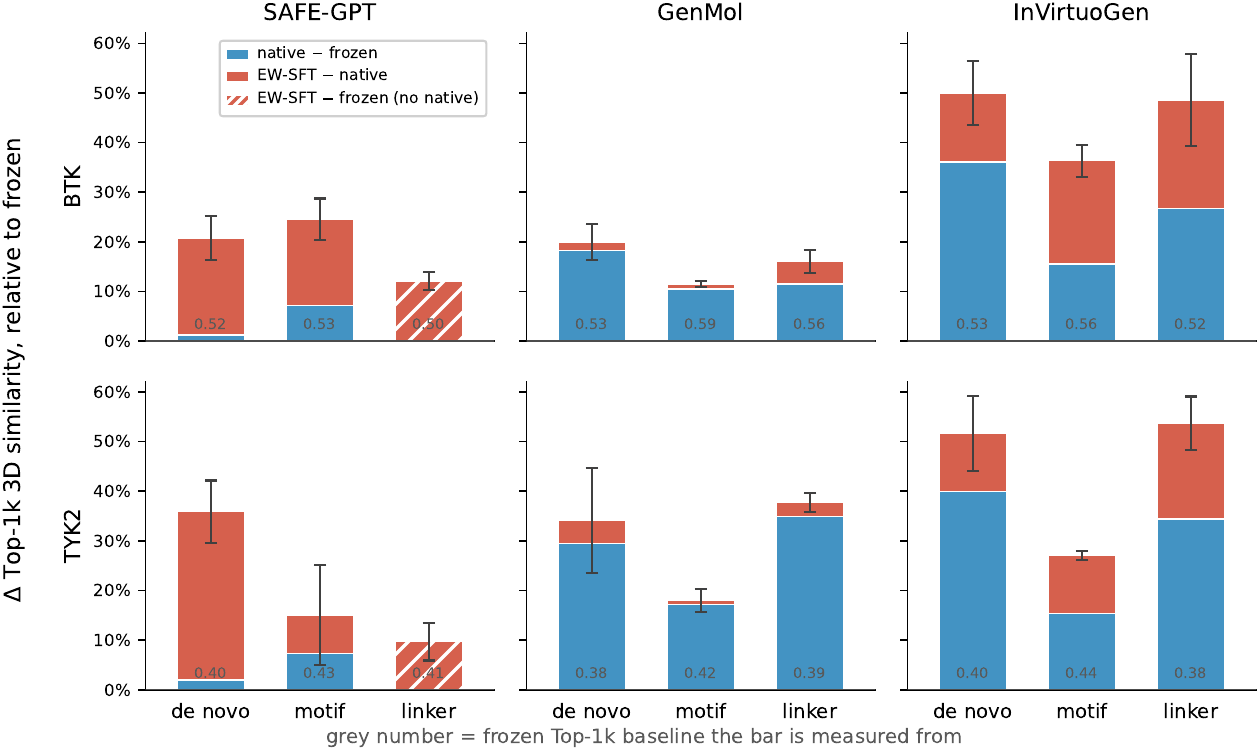}
\caption{\textbf{Relative Top-1k improvements over frozen sampling.}
Each bar shows the total improvement in Top-1k average score over frozen sampling (gain from native optimizer is blue, from EW-SFT is red), normalized by the frozen baseline (grey label). Hatched bars have no native term due to the incompatibility between PPO and SAFE-GPT's linker design process. Absolute gains are in Table~\ref{tab:gain}. EW-SFT provides an additional improvement over the native optimizer overall.}
\label{fig:gain}
\end{figure}

% \paragraph{EW-SFT adds headroom that the native optimizer leaves behind, on strong and weak baselines alike.}
\paragraph{Results.}
The native optimizers improve Top-1k average 3D shape similarity over frozen sampling, so goal-directed optimization is necessary even when the fixed fragments already supply some structural information (Figure~\ref{fig:gain} and Table~\ref{tab:gain}). EW-SFT adds significantly beyond that working optimizer across all settings,  with Holm-adjusted Wilcoxon signed rank test $p_{\text{SAFE-GPT}}=9.8\times10^{-4}$,
$p_{\text{GenMol}}=4.1\times10^{-2}$ and $p_{\text{InVirtuoGen}}=2.3\times10^{-5}$ (Appendix~\ref{app:stats}, Top-1k average in Table~\ref{tab:main-top1k} and AUC Top-10 in Table~\ref{tab:app-auc}). This is not merely a remedy for a weak baseline --- it reaches $+0.079$ over the strongest InVirtuoGen's Genetic-PPO, against $+0.091$ over the weakest SAFE-GPT's PPO. Neither is the gain concentrated on the easiest task, being nearly flat at $+0.064$ de novo, $+0.051$ motif, and $+0.055$ linker, nor does it follow from one model starting further behind, since the three priors sit within $0.012$ of each other on frozen Top-1k. The improvement is also not obtained by drifting toward the reference. EW-SFT is less 2D-similar to the reference than the native optimizer in $9$ of the $16$ cells while scoring higher in 3D in all $16$ (Table~\ref{tab:novelty2d}). Since the objective constrains shape alone, moving away from the reference in 2D while matching it more closely in 3D is consistent with satisfying the shape requirement using chemically distinct molecules rather than simply reconstructing the reference.

% ============================================================
% Table 1: Main Results — Top-1k (mean ± std, 3 seeds)
% ============================================================
\begin{table}[ht]
\caption{\textbf{Top-1k average scores of 3D shape similarity optimization.} \textbf{Bold} is the best value within each model block (REINVENT excluded). PPO is not evaluated for SAFE-GPT's linker design because its two-pass generation requires a joint policy over the complete generation trajectory.}
\label{tab:main-top1k}
\centering
\resizebox{\textwidth}{!}{
\begin{tabular}{ll cccccc |c}
\toprule
& & \multicolumn{3}{c}{BTK} & \multicolumn{3}{c}{TYK2} & \\
\cmidrule(lr){3-5} \cmidrule(lr){6-8}
Model & Method & De novo & Motif & Linker & De novo & Motif & Linker & Avg. \\
\midrule
\multirow{1}{*}{REINVENT}
  & REINFORCE & 0.530{\scriptsize$\pm$0.000} & 0.696{\scriptsize$\pm$0.010} & 0.576{\scriptsize$\pm$0.007} & 0.416{\scriptsize$\pm$0.001} & 0.494{\scriptsize$\pm$0.008} & 0.425{\scriptsize$\pm$0.003} & 0.523 \\
\midrule
\multirow{2}{*}{SAFE-GPT}
  & PPO & 0.528{\scriptsize$\pm$0.000} & 0.572{\scriptsize$\pm$0.012} & --- & 0.408{\scriptsize$\pm$0.001} & 0.459{\scriptsize$\pm$0.009} & --- & 0.492 \\
  & \oc EW-SFT (ours) & \oc \best{0.630}{\scriptsize$\pm$0.023} & \oc \best{0.665}{\scriptsize$\pm$0.022} & \oc \best{0.563}{\scriptsize$\pm$0.009} & \oc \best{0.544}{\scriptsize$\pm$0.025} & \oc \best{0.492}{\scriptsize$\pm$0.043} & \oc \best{0.446}{\scriptsize$\pm$0.014} & \oc \best{0.556} \\
\midrule
\multirow{2}{*}{GenMol}
  & GA & 0.622{\scriptsize$\pm$0.008} & 0.657{\scriptsize$\pm$0.007} & 0.624{\scriptsize$\pm$0.048} & 0.493{\scriptsize$\pm$0.009} & 0.487{\scriptsize$\pm$0.014} & 0.521{\scriptsize$\pm$0.004} & 0.567 \\
  & \oc EW-SFT (ours) & \oc \best{0.631}{\scriptsize$\pm$0.019} & \oc \best{0.663}{\scriptsize$\pm$0.003} & \oc \best{0.650}{\scriptsize$\pm$0.013} & \oc \best{0.510}{\scriptsize$\pm$0.040} & \oc \best{0.490}{\scriptsize$\pm$0.009} & \oc \best{0.532}{\scriptsize$\pm$0.007} & \oc \best{0.579} \\
\midrule
\multirow{2}{*}{InVirtuoGen}
  & Genetic-PPO & 0.723{\scriptsize$\pm$0.081} & 0.651{\scriptsize$\pm$0.009} & 0.656{\scriptsize$\pm$0.008} & 0.565{\scriptsize$\pm$0.012} & 0.507{\scriptsize$\pm$0.021} & 0.506{\scriptsize$\pm$0.041} & 0.602 \\
  & \oc EW-SFT (ours) & \oc \best{0.797}{\scriptsize$\pm$0.034} & \oc \best{0.769}{\scriptsize$\pm$0.017} & \oc \best{0.769}{\scriptsize$\pm$0.048} & \oc \best{0.612}{\scriptsize$\pm$0.030} & \oc \best{0.559}{\scriptsize$\pm$0.004} & \oc \best{0.578}{\scriptsize$\pm$0.020} & \oc \best{0.681} \\
\bottomrule
\end{tabular}
}
\end{table}

% ============================================================
% ============================================================
% Figure: gain decomposition (relative), generated by analysis/gain_figure.py --relative
% ============================================================

% ============================================================
\subsection{Ablation Study}
\label{sec:ablation}

\paragraph{Setup.} The mechanism of EW-SFT consists of three axes. \textbf{Source} determines whether training samples come from the genetic operator $G$ or directly from the pretrained model $p_\theta$. \textbf{Selection} determines whether fine-tuning uses the current round's batch or the elite buffer $E_t$. \textbf{Emphasis} is layered on top of selection and determines how strongly each selected sample contributes. Crossing the last two gives the EW-SFT variants in Table~\ref{tab:ablation-results}. Two settings are left out --- signed emphases on the elite buffer, which Proposition~\ref{prop:unbounded} rules ill-posed, and the batch-selection setting for InVirtuoGen, whose genetic population is already an elite buffer in the sense defined in Section~\ref{sec:ew-sft}. Table~\ref{tab:ablation-design} in Appendix~\ref{app:ablation-design} gives the resulting settings and the reasoning behind each exclusion.

% ============================================================
% Table: Ablation results
% ============================================================
\begin{table}[ht]
\caption{\textbf{Ablation results.} Top-1k mean$\pm$std 3D similarity over 3 runs at 20k oracle calls, for each model's native optimizer, the EW-SFT variants and the two standalone settings (GA only and SFT only). The best result in each task is marked in \textbf{bold}. SAFE-GPT PPO cannot run linker design. The last column is the mean difference from that model's native optimizer over its cells.}
\label{tab:ablation-results}
\centering
\resizebox{0.8\textwidth}{!}{
\begin{tabular}{llll cccc|c}
\toprule
& \multicolumn{3}{c}{Setting} & \multicolumn{2}{c}{BTK} & \multicolumn{2}{c}{TYK2} & \\
\cmidrule(lr){2-4} \cmidrule(lr){5-6} \cmidrule(lr){7-8}
Model & Source & Selection & Emphasis & De novo & Linker & De novo & Linker & $\overline{\Delta}_\text{var-nat}$ \\
\midrule
\multirow{9}{*}{SAFE-GPT}
  & \multicolumn{3}{l}{Native (PPO)} & 0.528{\scriptsize$\pm$0.000} & --- & 0.408{\scriptsize$\pm$0.001} & --- & --- \\
\cmidrule(l){2-4}
  & \multicolumn{8}{l}{EW-SFT (ours)} \\
  & \oc $G$ & \oc elite & \oc binary & \oc \best{0.630}{\scriptsize$\pm$0.023} & \oc 0.563{\scriptsize$\pm$0.009} & \oc \best{0.544}{\scriptsize$\pm$0.025} & \oc 0.446{\scriptsize$\pm$0.014} & \oc \best{$+$0.119} \\
  & \oc $G$ & \oc elite & \oc positive & \oc 0.593{\scriptsize$\pm$0.013} & \oc 0.563{\scriptsize$\pm$0.011} & \oc 0.507{\scriptsize$\pm$0.031} & \oc 0.455{\scriptsize$\pm$0.014} & \oc $+$0.082 \\
  & \oc $G$ & \oc batch & \oc positive & \oc 0.620{\scriptsize$\pm$0.029} & \oc 0.559{\scriptsize$\pm$0.005} & \oc 0.491{\scriptsize$\pm$0.011} & \oc \best{0.459}{\scriptsize$\pm$0.027} & \oc $+$0.088 \\
  & \oc $G$ & \oc batch & \oc signed & \oc 0.616{\scriptsize$\pm$0.023} & \oc \best{0.566}{\scriptsize$\pm$0.002} & \oc 0.496{\scriptsize$\pm$0.027} & \oc 0.458{\scriptsize$\pm$0.011} & \oc $+$0.088 \\
\cmidrule(l){2-4}
  & \multicolumn{8}{l}{Standalone} \\
  & $G$ & --- & --- & 0.607{\scriptsize$\pm$0.029} & 0.564{\scriptsize$\pm$0.005} & 0.501{\scriptsize$\pm$0.008} & 0.438{\scriptsize$\pm$0.008} & $+$0.086 \\
  & $p_\theta$ & elite & binary & 0.524{\scriptsize$\pm$0.001} & 0.553{\scriptsize$\pm$0.005} & 0.399{\scriptsize$\pm$0.001} & 0.421{\scriptsize$\pm$0.002} & $-$0.006 \\
\midrule
\multirow{9}{*}{GenMol}
  & \multicolumn{3}{l}{Native (GA)} & 0.622{\scriptsize$\pm$0.008} & 0.624{\scriptsize$\pm$0.048} & 0.493{\scriptsize$\pm$0.009} & 0.521{\scriptsize$\pm$0.004} & --- \\
\cmidrule(l){2-4}
  & \multicolumn{8}{l}{EW-SFT (ours)} \\
  & \oc $G$ & \oc elite & \oc binary & \oc 0.631{\scriptsize$\pm$0.019} & \oc 0.650{\scriptsize$\pm$0.013} & \oc 0.510{\scriptsize$\pm$0.040} & \oc \best{0.532}{\scriptsize$\pm$0.007} & \oc $+$0.016 \\
  & \oc $G$ & \oc elite & \oc positive & \oc 0.638{\scriptsize$\pm$0.018} & \oc \best{0.664}{\scriptsize$\pm$0.014} & \oc 0.505{\scriptsize$\pm$0.029} & \oc 0.529{\scriptsize$\pm$0.007} & \oc \best{$+$0.019} \\
  & \oc $G$ & \oc batch & \oc positive & \oc 0.631{\scriptsize$\pm$0.020} & \oc 0.635{\scriptsize$\pm$0.045} & \oc \best{0.513}{\scriptsize$\pm$0.033} & \oc 0.520{\scriptsize$\pm$0.023} & \oc $+$0.010 \\
  & \oc $G$ & \oc batch & \oc signed & \oc \best{0.644}{\scriptsize$\pm$0.007} & \oc 0.635{\scriptsize$\pm$0.025} & \oc 0.513{\scriptsize$\pm$0.027} & \oc 0.516{\scriptsize$\pm$0.014} & \oc $+$0.012 \\
\cmidrule(l){2-4}
  & \multicolumn{8}{l}{Standalone} \\
  & $G$ & --- & --- & 0.622{\scriptsize$\pm$0.008} & 0.624{\scriptsize$\pm$0.048} & 0.493{\scriptsize$\pm$0.009} & 0.521{\scriptsize$\pm$0.004} & $+$0.000 \\
  & $p_\theta$ & elite & binary & 0.531{\scriptsize$\pm$0.000} & 0.520{\scriptsize$\pm$0.001} & 0.385{\scriptsize$\pm$0.000} & 0.405{\scriptsize$\pm$0.002} & $-$0.105 \\
\midrule
\multirow{8}{*}{InVirtuoGen}
  & \multicolumn{3}{l}{Native (Genetic-PPO)} & 0.723{\scriptsize$\pm$0.081} & 0.656{\scriptsize$\pm$0.008} & 0.565{\scriptsize$\pm$0.012} & 0.506{\scriptsize$\pm$0.041} & --- \\
\cmidrule(l){2-4}
  & \multicolumn{8}{l}{EW-SFT (ours)} \\
  & \oc $G$ & \oc elite & \oc binary & \oc \best{0.797}{\scriptsize$\pm$0.034} & \oc \best{0.769}{\scriptsize$\pm$0.048} & \oc \best{0.612}{\scriptsize$\pm$0.030} & \oc \best{0.578}{\scriptsize$\pm$0.020} & \oc \best{$+$0.076} \\
  & \oc $G$ & \oc elite & \oc positive & \oc 0.778{\scriptsize$\pm$0.023} & \oc 0.725{\scriptsize$\pm$0.008} & \oc 0.597{\scriptsize$\pm$0.021} & \oc 0.547{\scriptsize$\pm$0.039} & \oc $+$0.049 \\
  & \oc $G$ & \oc elite & \oc signed & \oc 0.710{\scriptsize$\pm$0.039} & \oc 0.661{\scriptsize$\pm$0.048} & \oc 0.543{\scriptsize$\pm$0.048} & \oc 0.550{\scriptsize$\pm$0.037} & \oc $+$0.003 \\
\cmidrule(l){2-4}
  & \multicolumn{8}{l}{Standalone} \\
  & $G$ & --- & --- & 0.749{\scriptsize$\pm$0.025} & 0.700{\scriptsize$\pm$0.058} & 0.576{\scriptsize$\pm$0.050} & 0.556{\scriptsize$\pm$0.057} & $+$0.032 \\
  & $p_\theta$ & elite & binary & 0.532{\scriptsize$\pm$0.001} & 0.565{\scriptsize$\pm$0.003} & 0.409{\scriptsize$\pm$0.002} & 0.411{\scriptsize$\pm$0.003} & $-$0.134 \\
\bottomrule
\end{tabular}
}
\end{table}

% \paragraph{The source and quality of the samples matter more than their emphasis.}
\paragraph{Source of training samples.}
In two standalone cases, capturing training samples from the genetic operator alone improves on the native optimizer, which aligns with the claim that genetic algorithm is already a strong baseline in this domain \citep{graphGA, tripp2023geneticalgorithmsstrongbaselines}. Directly using the policy-generated samples fails to enhance the model. SFT-only holds the elite buffer and the binary emphasis fixed, it is therefore essentially distillation. In our setting, the generator sharpens around elites drawn from its own output but does not reliably introduce improvements beyond its prior. The genetic operator is what breaks that closed loop by mutating or refining the samples for EW-SFT to further update.

% \paragraph{Elite selection alone carries the reward signal.}
\paragraph{Reward guidance.}
Given that source, elite selection into a rolling buffer is sufficient to deliver the reward to the update. As a molecule enters the buffer only by outscoring the minimum there, the elites are the best seen so far and intrinsically carry the reward signal, so a continuous emphasis adds nothing that further distinguishes among them. When fine-tuning uses the full current batch without elite filtering, a positive emphasis does improve the policy model and mildly surpasses the GA alone, because the emphasis is then the only channel through which the reward can act. Applying a positive emphasis on top of the elite buffer is redundant and weakens the update by only using the positive-weighted elites. The signed emphasis reduces $0.073$ average improvement on InVirtuoGen against the same source under a binary emphasis and has the diverging issue as shown in Proposition~\ref{prop:unbounded} (Appendix~\ref{app:signed-emph}). Overall, neither the source nor the selection component suffices alone. We report EW-SFT with the binary emphasis over the elite buffer based on the genetic operator throughout.

\begin{figure}[ht]
\centering
\includegraphics[width=0.9\textwidth]{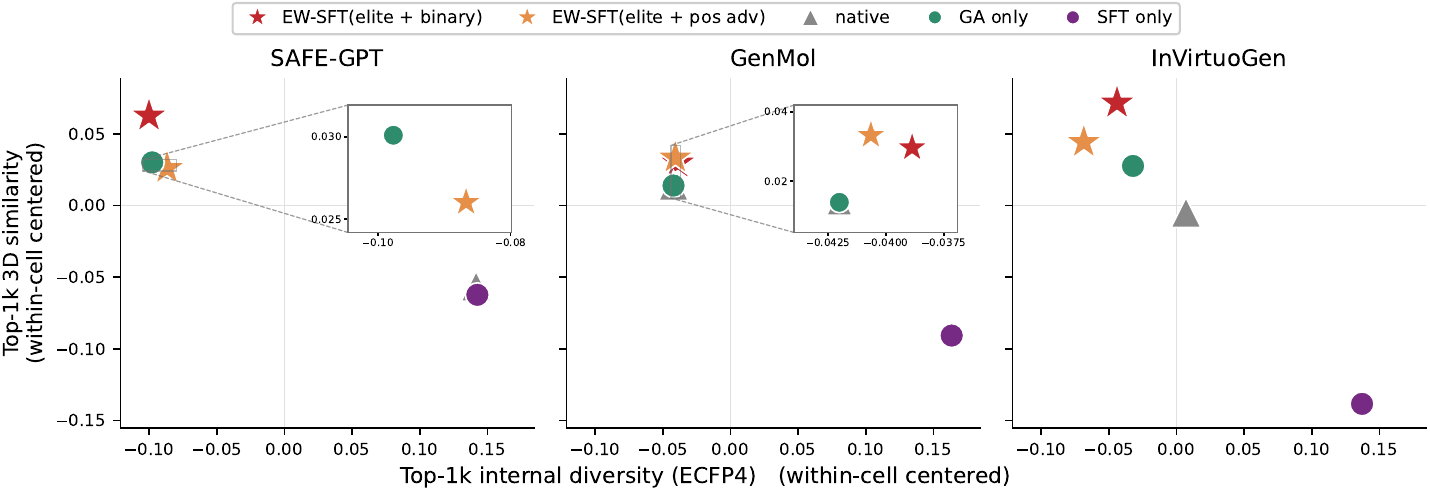}
\caption{\textbf{EW-SFT gains score while staying diverse.} Each point is a model's one ablation setting, averaged over the task cells in which every setting ran (SAFE-GPT's linker cells are excluded). Both axes are centered within a model. GenMol's native optimizer is its GA, so those two markers coincide. EW-SFT improves over the native optimizer on score while staying close to the genetic algorithm on diversity.}
\label{fig:qd}
\end{figure}

% \paragraph{EW-SFT keeps improving from a diverse pool.}
\paragraph{Quality and diversity tradeoff.}
Figure~\ref{fig:qd} exhibits the quality and diversity tradeoff of the Top-1k samples across each ablation setting. Higher-scoring settings tend to be less diverse (per-cell Pearson $r$ averaging $-0.52$, negative in $11$ of $12$ cells, $p=5\times10^{-4}$), but EW-SFT with elite+binary is the only setting that sits significantly above what that tradeoff predicts ($p=4\times10^{-4}$), while the native optimizer is significantly below it. Specifically, Figure~\ref{fig:exploit-explore} shows that the moving-average batch diversity under EW-SFT falls early and then settles onto a plateau rather than continuing to decline, holding near $0.65$--$0.68$ on SAFE-GPT and GenMol and $0.77$--$0.79$ on InVirtuoGen. The batch maximum keeps improving over that same stretch on SAFE-GPT TYK2 and InVirtuoGen BTK de novo design, where the native optimizer has either flattened or stopped under its own novelty rule at roughly $8$k calls (Figures~\ref{fig:safe-convergence}--\ref{fig:ivg-convergence}). A search whose diversity levels off while its best molecules keep improving is consistent with continued exploration rather than complete collapse.

% ============================================================
% Figure: quality-diversity by model. Generated by
%   python analysis/qd_bymodel.py --metric intdiv --legend-in-panel \
%          --include-sft-only --out figures/qd_bymodel_intdiv_withSFT
% ============================================================

\subsection{Generalization to Held-out References and PMO}
\label{sec:generalizability}

\paragraph{Fragment-constrained benchmark.}
We select four reference drugs (Table~\ref{tab:anchors}) from the fragment-constrained benchmark \citep{safe-gpt}, and use 2D ECFP4 Tanimoto similarity~\citep{rogers2010extended} to the reference compound as the oracle. We first follow the common practice of fragment-constrained evaluation by frozen sampling with the best backbone, InVirtuoGen, under fragment anchor conditions \citep{safe-gpt,lee2025genmol,kaech2026refine}. To our knowledge, we are the first to adapt the native de novo optimizer to this benchmark's reference compounds and find that it substantially improves the average performance by $+0.135$ (adjusted
$p_{\text{nat-frz}}=2.4\times10^{-7}$) (Table~\ref{tab:generalizability}). Our best EW-SFT
arm (genetic source, elite selection, and binary emphasis) further improves $+0.086$ on
top, winning $7$ of $8$ tasks (adjusted $p_{\text{ours-nat}}=6.4\times10^{-5}$).

% ============================================================
% Generalizability: held-out references x held-out objective
% Body generated by analysis/generalizability_table.py
% ============================================================
\begin{table}[ht]
\caption{\textbf{Top-1k average scores of 2D ECFP similarity on the held-out references.} The results are mean$\pm$std over 3 runs at 20k oracle calls. The best result for each task is marked in bold. The four drug compounds and their fragment anchors are in Table~\ref{tab:anchors}.}
\label{tab:generalizability}
\centering
\resizebox{\textwidth}{!}{
\begin{tabular}{lcccccccc|c}
\toprule
& \multicolumn{2}{c}{Eliglustat} & \multicolumn{2}{c}{Erlotinib} & \multicolumn{2}{c}{Futibatinib} & \multicolumn{2}{c}{Spirapril} & \\
\cmidrule(lr){2-3} \cmidrule(lr){4-5} \cmidrule(lr){6-7} \cmidrule(lr){8-9}
Method & motif & linker & motif & linker & motif & linker & motif & linker & Avg. \\
\midrule
  Frozen (Top-1k) & 0.372{\scriptsize$\pm$0.001} & 0.433{\scriptsize$\pm$0.002} & 0.263{\scriptsize$\pm$0.001} & 0.331{\scriptsize$\pm$0.001} & 0.331{\scriptsize$\pm$0.000} & 0.425{\scriptsize$\pm$0.001} & 0.483{\scriptsize$\pm$0.001} & 0.522{\scriptsize$\pm$0.002} & 0.395 \\
  Genetic-PPO & \best{0.611}{\scriptsize$\pm$0.021} & 0.645{\scriptsize$\pm$0.005} & 0.411{\scriptsize$\pm$0.060} & 0.448{\scriptsize$\pm$0.008} & 0.441{\scriptsize$\pm$0.040} & 0.491{\scriptsize$\pm$0.013} & 0.562{\scriptsize$\pm$0.016} & 0.630{\scriptsize$\pm$0.009} & 0.530 \\
  \oc EW-SFT & \oc 0.588{\scriptsize$\pm$0.066} & \oc \best{0.740}{\scriptsize$\pm$0.091} & \oc \best{0.453}{\scriptsize$\pm$0.085} & \oc \best{0.540}{\scriptsize$\pm$0.016} & \oc \best{0.520}{\scriptsize$\pm$0.057} & \oc \best{0.605}{\scriptsize$\pm$0.072} & \oc \best{0.743}{\scriptsize$\pm$0.033} & \oc \best{0.743}{\scriptsize$\pm$0.056} & \oc \best{0.616} \\
\bottomrule
\end{tabular}
}
\end{table}

\paragraph{Practical Molecular Optimization.}
PMO~\citep{gao2022sample} evaluates sample-efficient de novo optimization under a $10$k oracle-call budget. We further evaluate elite$+$binary EW-SFT with InVirtuoGen. Table~\ref{tab:pmo} shows comparable aggregate performance to Genetic-PPO under both prescreening ($18.863$ vs.\ $18.992$, adjusted $p=0.69$) and cold-start ($16.385$ vs.\ $16.674$, adjusted $p=0.87$). Paired comparisons do not detect a significant difference in either setting.  Genetic-PPO uses stored behavior-policy generation-trajectory log-probabilities, importance ratios, ratio clipping, negative advantage scaling, and experience replay. In contrast, EW-SFT uses a native-loss step on the elite set and requires no generation-trajectory log-probability. Per-oracle, EW-SFT takes the best result on $10$ of the $23$ prescreening oracles against Genetic-PPO's $9$, and on $8$ against its $10$ under cold-start.

% ============================================================
% PMO: 23 TDC oracles, both initialisation settings
% ============================================================
\begin{table}[ht]
\caption{\textbf{PMO benchmark.} AUC Top-10 scores (mean$\pm$std over 3 runs) under the two initialization settings used by InVirtuoGen, with a $10$k oracle-call budget per run. \emph{With prescreening} initializes the population with the $40$ highest-scoring ZINC250k molecules under the task oracle; the $250\mathrm{k}\times23$ prescreening evaluations are excluded from this budget. \emph{Cold-start} initializes from the generator alone. \textbf{Bold} indicates the best mean within each setting.}
\label{tab:pmo}
\centering
\resizebox{\textwidth}{!}{
\begin{tabular}{lcccc|ccccc}
\toprule
& \multicolumn{4}{c}{w/ Prescreening} & \multicolumn{5}{c}{Cold-Start (w/o Prescreening)} \\
\cmidrule(lr){2-5} \cmidrule(lr){6-10}
Oracle & \oc \shortstack{InVirtuoGen\\ EW-SFT (ours)} & \shortstack{InVirtuoGen\\ Genetic-PPO} & GenMol & f-RAG
       & \oc \shortstack{InVirtuoGen\\ EW-SFT (ours)} & \shortstack{InVirtuoGen\\ Genetic-PPO} & Gen. GFN & Mol GA & REINVENT \\
\midrule
  albuterol\_similarity & \oc 0.973{\scriptsize$\pm$0.015} & 0.975{\scriptsize$\pm$0.016} & 0.937{\scriptsize$\pm$0.010} & \best{0.977}{\scriptsize$\pm$0.002} & \oc \best{0.966}{\scriptsize$\pm$0.011} & 0.950{\scriptsize$\pm$0.017} & 0.949{\scriptsize$\pm$0.010} & 0.896{\scriptsize$\pm$0.035} & 0.882{\scriptsize$\pm$0.006} \\
  amlodipine\_mpo & \oc 0.812{\scriptsize$\pm$0.023} & \best{0.836}{\scriptsize$\pm$0.031} & 0.810{\scriptsize$\pm$0.012} & 0.749{\scriptsize$\pm$0.019} & \oc 0.740{\scriptsize$\pm$0.089} & 0.733{\scriptsize$\pm$0.043} & \best{0.761}{\scriptsize$\pm$0.019} & 0.688{\scriptsize$\pm$0.039} & 0.635{\scriptsize$\pm$0.035} \\
  celecoxib\_rediscovery & \oc 0.832{\scriptsize$\pm$0.018} & \best{0.839}{\scriptsize$\pm$0.013} & 0.826{\scriptsize$\pm$0.018} & 0.778{\scriptsize$\pm$0.007} & \oc \best{0.819}{\scriptsize$\pm$0.026} & 0.798{\scriptsize$\pm$0.028} & 0.802{\scriptsize$\pm$0.029} & 0.567{\scriptsize$\pm$0.083} & 0.713{\scriptsize$\pm$0.067} \\
  deco\_hop & \oc \best{0.981}{\scriptsize$\pm$0.007} & 0.968{\scriptsize$\pm$0.012} & 0.960{\scriptsize$\pm$0.010} & 0.936{\scriptsize$\pm$0.011} & \oc 0.714{\scriptsize$\pm$0.068} & \best{0.748}{\scriptsize$\pm$0.109} & 0.733{\scriptsize$\pm$0.109} & 0.649{\scriptsize$\pm$0.025} & 0.666{\scriptsize$\pm$0.044} \\
  drd2 & \oc \best{0.995}{\scriptsize$\pm$0.000} & \best{0.995}{\scriptsize$\pm$0.000} & \best{0.995}{\scriptsize$\pm$0.000} & 0.992{\scriptsize$\pm$0.000} & \oc 0.984{\scriptsize$\pm$0.001} & \best{0.985}{\scriptsize$\pm$0.002} & 0.974{\scriptsize$\pm$0.006} & 0.936{\scriptsize$\pm$0.016} & 0.945{\scriptsize$\pm$0.007} \\
  fexofenadine\_mpo & \oc 0.897{\scriptsize$\pm$0.035} & \best{0.904}{\scriptsize$\pm$0.000} & 0.894{\scriptsize$\pm$0.028} & 0.856{\scriptsize$\pm$0.016} & \oc \best{0.863}{\scriptsize$\pm$0.031} & 0.845{\scriptsize$\pm$0.016} & 0.856{\scriptsize$\pm$0.039} & 0.825{\scriptsize$\pm$0.019} & 0.784{\scriptsize$\pm$0.006} \\
  gsk3b & \oc \best{0.989}{\scriptsize$\pm$0.002} & 0.988{\scriptsize$\pm$0.001} & 0.986{\scriptsize$\pm$0.003} & 0.969{\scriptsize$\pm$0.003} & \oc 0.914{\scriptsize$\pm$0.045} & \best{0.952}{\scriptsize$\pm$0.016} & 0.881{\scriptsize$\pm$0.042} & 0.843{\scriptsize$\pm$0.039} & 0.865{\scriptsize$\pm$0.043} \\
  isomers\_c7h8n2o2 & \oc \best{0.991}{\scriptsize$\pm$0.001} & 0.988{\scriptsize$\pm$0.002} & 0.942{\scriptsize$\pm$0.004} & 0.955{\scriptsize$\pm$0.008} & \oc 0.964{\scriptsize$\pm$0.013} & 0.968{\scriptsize$\pm$0.005} & \best{0.969}{\scriptsize$\pm$0.003} & 0.878{\scriptsize$\pm$0.026} & 0.852{\scriptsize$\pm$0.036} \\
  isomers\_c9h10n2o2pf2cl & \oc 0.892{\scriptsize$\pm$0.030} & \best{0.898}{\scriptsize$\pm$0.018} & 0.833{\scriptsize$\pm$0.014} & 0.850{\scriptsize$\pm$0.005} & \oc 0.878{\scriptsize$\pm$0.025} & 0.874{\scriptsize$\pm$0.013} & \best{0.897}{\scriptsize$\pm$0.007} & 0.865{\scriptsize$\pm$0.012} & 0.642{\scriptsize$\pm$0.054} \\
  jnk3 & \oc 0.900{\scriptsize$\pm$0.044} & 0.898{\scriptsize$\pm$0.031} & \best{0.906}{\scriptsize$\pm$0.023} & 0.904{\scriptsize$\pm$0.004} & \oc \best{0.841}{\scriptsize$\pm$0.021} & 0.825{\scriptsize$\pm$0.016} & 0.764{\scriptsize$\pm$0.069} & 0.702{\scriptsize$\pm$0.123} & 0.783{\scriptsize$\pm$0.023} \\
  median1 & \oc 0.385{\scriptsize$\pm$0.011} & 0.386{\scriptsize$\pm$0.003} & \best{0.398}{\scriptsize$\pm$0.000} & 0.340{\scriptsize$\pm$0.007} & \oc 0.354{\scriptsize$\pm$0.016} & 0.342{\scriptsize$\pm$0.008} & \best{0.379}{\scriptsize$\pm$0.010} & 0.257{\scriptsize$\pm$0.009} & 0.356{\scriptsize$\pm$0.009} \\
  median2 & \oc 0.376{\scriptsize$\pm$0.008} & \best{0.377}{\scriptsize$\pm$0.006} & 0.359{\scriptsize$\pm$0.004} & 0.323{\scriptsize$\pm$0.005} & \oc \best{0.310}{\scriptsize$\pm$0.013} & 0.288{\scriptsize$\pm$0.008} & 0.294{\scriptsize$\pm$0.007} & 0.301{\scriptsize$\pm$0.021} & 0.276{\scriptsize$\pm$0.008} \\
  mestranol\_similarity & \oc 0.987{\scriptsize$\pm$0.008} & \best{0.991}{\scriptsize$\pm$0.002} & 0.982{\scriptsize$\pm$0.000} & 0.671{\scriptsize$\pm$0.021} & \oc \best{0.830}{\scriptsize$\pm$0.053} & 0.797{\scriptsize$\pm$0.033} & 0.708{\scriptsize$\pm$0.057} & 0.591{\scriptsize$\pm$0.053} & 0.618{\scriptsize$\pm$0.048} \\
  osimertinib\_mpo & \oc \best{0.882}{\scriptsize$\pm$0.017} & 0.881{\scriptsize$\pm$0.012} & 0.876{\scriptsize$\pm$0.008} & 0.866{\scriptsize$\pm$0.009} & \oc 0.859{\scriptsize$\pm$0.025} & \best{0.870}{\scriptsize$\pm$0.005} & 0.860{\scriptsize$\pm$0.008} & 0.844{\scriptsize$\pm$0.015} & 0.837{\scriptsize$\pm$0.009} \\
  perindopril\_mpo & \oc \best{0.758}{\scriptsize$\pm$0.012} & 0.753{\scriptsize$\pm$0.019} & 0.718{\scriptsize$\pm$0.012} & 0.681{\scriptsize$\pm$0.017} & \oc 0.619{\scriptsize$\pm$0.035} & \best{0.645}{\scriptsize$\pm$0.032} & 0.595{\scriptsize$\pm$0.014} & 0.547{\scriptsize$\pm$0.022} & 0.537{\scriptsize$\pm$0.016} \\
  qed & \oc \best{0.944}{\scriptsize$\pm$0.000} & 0.943{\scriptsize$\pm$0.000} & 0.942{\scriptsize$\pm$0.000} & 0.939{\scriptsize$\pm$0.001} & \oc \best{0.942}{\scriptsize$\pm$0.001} & \best{0.942}{\scriptsize$\pm$0.000} & \best{0.942}{\scriptsize$\pm$0.000} & 0.941{\scriptsize$\pm$0.001} & 0.941{\scriptsize$\pm$0.000} \\
  ranolazine\_mpo & \oc \best{0.858}{\scriptsize$\pm$0.014} & 0.854{\scriptsize$\pm$0.012} & 0.821{\scriptsize$\pm$0.011} & 0.820{\scriptsize$\pm$0.016} & \oc \best{0.852}{\scriptsize$\pm$0.004} & 0.848{\scriptsize$\pm$0.010} & 0.819{\scriptsize$\pm$0.018} & 0.804{\scriptsize$\pm$0.011} & 0.760{\scriptsize$\pm$0.009} \\
  scaffold\_hop & \oc 0.652{\scriptsize$\pm$0.011} & \best{0.711}{\scriptsize$\pm$0.081} & 0.628{\scriptsize$\pm$0.008} & 0.576{\scriptsize$\pm$0.014} & \oc 0.597{\scriptsize$\pm$0.043} & 0.589{\scriptsize$\pm$0.021} & \best{0.615}{\scriptsize$\pm$0.100} & 0.527{\scriptsize$\pm$0.025} & 0.560{\scriptsize$\pm$0.019} \\
  sitagliptin\_mpo & \oc \best{0.763}{\scriptsize$\pm$0.036} & 0.743{\scriptsize$\pm$0.022} & 0.584{\scriptsize$\pm$0.034} & 0.601{\scriptsize$\pm$0.011} & \oc 0.624{\scriptsize$\pm$0.055} & \best{0.709}{\scriptsize$\pm$0.029} & 0.634{\scriptsize$\pm$0.039} & 0.582{\scriptsize$\pm$0.040} & 0.021{\scriptsize$\pm$0.003} \\
  thiothixene\_rediscovery & \oc 0.629{\scriptsize$\pm$0.011} & 0.652{\scriptsize$\pm$0.024} & \best{0.692}{\scriptsize$\pm$0.123} & 0.584{\scriptsize$\pm$0.009} & \oc 0.621{\scriptsize$\pm$0.048} & \best{0.625}{\scriptsize$\pm$0.014} & 0.583{\scriptsize$\pm$0.034} & 0.519{\scriptsize$\pm$0.041} & 0.534{\scriptsize$\pm$0.013} \\
  troglitazone\_rediscovery & \oc 0.850{\scriptsize$\pm$0.015} & 0.853{\scriptsize$\pm$0.003} & \best{0.867}{\scriptsize$\pm$0.022} & 0.448{\scriptsize$\pm$0.017} & \oc 0.554{\scriptsize$\pm$0.052} & \best{0.595}{\scriptsize$\pm$0.053} & 0.511{\scriptsize$\pm$0.054} & 0.427{\scriptsize$\pm$0.031} & 0.441{\scriptsize$\pm$0.032} \\
  valsartan\_smarts & \oc 0.886{\scriptsize$\pm$0.010} & \best{0.935}{\scriptsize$\pm$0.012} & 0.822{\scriptsize$\pm$0.042} & 0.627{\scriptsize$\pm$0.058} & \oc 0.000{\scriptsize$\pm$0.000} & \best{0.210}{\scriptsize$\pm$0.297} & 0.135{\scriptsize$\pm$0.271} & 0.000{\scriptsize$\pm$0.000} & 0.178{\scriptsize$\pm$0.358} \\
  zaleplon\_mpo & \oc \best{0.632}{\scriptsize$\pm$0.008} & 0.624{\scriptsize$\pm$0.040} & 0.584{\scriptsize$\pm$0.011} & 0.486{\scriptsize$\pm$0.004} & \oc 0.539{\scriptsize$\pm$0.006} & 0.536{\scriptsize$\pm$0.006} & \best{0.552}{\scriptsize$\pm$0.033} & 0.519{\scriptsize$\pm$0.029} & 0.358{\scriptsize$\pm$0.062} \\
  \midrule
  Sum & \oc 18.863 & \textbf{18.992} & 18.362 & 16.928 & \oc 16.385 & \textbf{16.674} & 16.213 & 14.708 & 14.184 \\
\bottomrule
\end{tabular}}
\end{table}

\section{Conclusion}

Goal-directed molecular optimization often relies on policy-gradient RL, whose generation-trajectory log-probabilities depend on how a particular generator and generation procedure produce a molecule. We show that oracle feedback can instead guide native-loss fine-tuning through selection: genetically evolved, high-scoring molecules are retained in a rolling elite buffer, and the generator is updated on them using its own pretraining loss.
EW-SFT is a unified optimizer that adapts to each generator through its native proposal mechanism and pretraining loss. Across autoregressive, masked-diffusion, and discrete-flow generators, and across de novo, motif-extension, and linker-design tasks, it significantly improves on the native optimizer's Top-1k 3D shape similarity, generalizes to held-out references under a 2D oracle, and achieves comparable performance to Genetic-PPO on PMO. Across three architectures, three tasks, six references, both 3D and 2D objectives, and $23$ further oracles on an external benchmark, we demonstrate that EW-SFT is a general and effective method for goal-directed molecular optimization.

% this part is more like a limitation discussion and futher work, not conclusion. 
% However, although we compute multiple conformers for the 3D shape oracle, a missed pose could score a good molecule poorly, and there is a balance between the enumeration and computational cost. In the future, EW-SFT can be further used to optimize molecular design towards docking scores or predicted binding affinities, whose landscapes are rougher. The three generators we optimize use fragment-sequence representations, whereas a genetic operator is equally definable on graphs. Since our update requires only a scored sample and the model's own training loss, it should carry over to graph-based and future generalist molecular models, and to goal-directed optimization in other fields, bypassing a full reinforcement learning implementation.

\newpage
% \section*{AI Use Statement}

% In this work, we used generative AI tools to improve the readability and grammar of the manuscript, and to assist with debugging implementation scripts. We did not use generative AI tools to generate datasets, develop the conceptual framework, or design the experimental methodology. We have reviewed all AI-assisted work. We take responsibility for the final content of this work, including text, claims or artifacts produced with the aid of generative AI.

\section*{Ethics Statement}

EW-SFT is intended to improve computational molecular design. Generated molecules should be evaluated under appropriate institutional, safety, and regulatory oversight. Improvements under 3D-shape and 2D-similarity oracles are computational evidence of optimization quality and do not by themselves establish binding, selectivity, synthesis feasibility, safety, or biological activity. Any downstream use should therefore include appropriate experimental validation before biological or therapeutic claims are made.

\section*{Reproducibility Statement}

To reproduce the results, Appendix~\ref{app:impl} details the generators' checkpoint version, the design and hyperparameter settings of genetic operators, optimizers, and the oracle. Appendix~\ref{app:runtime} shows the computational resources and runtime for our experiments. All the data and source code will be open-sourced after peer-review.

% \section*{Author Contributions}

% \section*{Acknowledgments}

\bibliography{iclr2027_conference}
\bibliographystyle{iclr2027_conference}

% ============================================================
\clearpage
\appendix

%%%%%%%%%%%%%
\section{Related Work}
\label{sec:related}

\paragraph{Recent molecular generative models.} The REINVENT family \citep{Olivecrona2017,reinvent4} established the recurrent Simplified Molecular Input Line Entry System (SMILES) \citep{doi:10.1021/ci00057a005} decoder trained with policy gradients for goal-directed design, and its per-task variants LibINVENT \citep{libinvent} and Link-INVENT \citep{linkinvent} remain the reference implementations for constrained generation. Later work moved toward a single generalist model. SAFE-GPT \citep{safe-gpt} introduced the Sequential Attachment-based Fragment Embedding (SAFE), a fragment-based representation that enables universal molecular design, and trained a GPT-2 backbone on it. GenMol \citep{lee2025genmol} applied a masked diffusion language model \citep{austin2023structureddenoisingdiffusionmodels,sahoo2024simpleeffectivemaskeddiffusion} on the same representation, decoupling next-token prediction from SAFE's order-agnostic structure. InVirtuoGen \citep{kaech2026refine} argues that generation is refinement rather than completion, and learns a discrete flow model \citep{campbell2024generativeflowsdiscretestatespaces} transporting a uniform source over the vocabulary to the data distribution. Graph-GRPO \citep{zhu2026graphgrpotraininggraphflow} works on graphs instead of sequences, deriving analytic transition probabilities for graph flow models to form PPO-style~\citep{ppo} importance ratios, illustrating the architecture-specific policy formulation required by policy-gradient RL.

\paragraph{Tasks and evaluations in molecular generation.} Frozen sampling is typically used as the first evaluation step for de novo and fragment-constrained generations, where samples generated by the pretrained model are scored for validity, uniqueness, diversity, and quality. SAFE \citep{safe-gpt} collected ten existing drugs and built a benchmark for fragment-constrained frozen sampling. Another task is goal-directed optimization, where GuacaMol \citep{GuacaMol} first standardized the objectives and PMO \citep{gao2022sample} fixed $10$k calls over $23$ oracles. Each architecture used in this paper~\citep{safe-gpt,lee2025genmol,kaech2026refine} has been tuned in this regime. However, Graph-GRPO's oracle calls spent on fine-tuning each task are not counted, and the oracle budget applies only to the evaluation run that follows. Although these models achieved SOTA scores, the tasks are de novo, so whether their optimizers are effective under fragment constraints is untested.

Optimization under a fixed substructure has been done before. LibINVENT \citep{libinvent} decorates a fixed scaffold with an RL-trained recurrent decorator under a 3D shape objective, and Link-INVENT \citep{linkinvent} generates
linkers between fixed warheads with docking in the loop. DRlinker \citep{drlinker} rewards linker design for
high 3D similarity at low 2D similarity with a transformer sequence-to-sequence policy. MoLeR \citep{moler} decodes a molecular graph motif by motif and optimizes by latent-space search, and FREED \citep{freed} attaches fragments to a growing graph under PPO. \cite{rush} argues that confining generation to a predefined substructure limits the exploration of the chemical space. These are per-task architectures with specific optimizers, whether one update rule can apply to all of them is the question this paper takes up.

\paragraph{Likelihood updates as a substitute for policy gradients.} Many molecular optimizations use policy-gradient RL, which needs an architecture-specific adaptation. Many studies keep that update stable by adding machinery around it: a prior-likelihood anchor~\citep{Olivecrona2017}, diversity filters and replay memory~\citep{reinvent4}, an augmented likelihood~\citep{Thomas2022}, and experience replay with SMILES augmentation~\citep{10.1021/jacsau.4c00066}. Each component improves sample efficiency but adds its own hyperparameters to tune~\citep{gao2022sample,10.1021/acs.jcim.5c02053}. We take the opposite direction, removing the trajectory-level formulation rather than stabilizing it. Advantage-weighted regression \citep{peng2019advantageweightedregressionsimplescalable} turns the reward into a per-sample weight and fits the model to its own samples by weighted maximum likelihood, bypassing the importance ratio entirely. But the exponentiated advantage has a temperature to tune, and it needs a value function estimated alongside the policy. Advantage-weighted matching \citep{xue2025advantageweightedmatchingaligning} removes the value function by weighting a diffusion model's own pretraining loss, on continuous latents with an advantage weight. In the molecular domain, hill-climbing selects the top-scoring samples a model has generated and refits the model to them by maximum likelihood~\citep{neil2018exploring}, and GEGL~\citep{ahn2020guidingdeepmolecularoptimization} adds an external graph-based genetic expert improvement that mutates the contents of a max-reward queue, training an LSTM to imitate that queue by unweighted maximum likelihood. Both take the reward through selection rather than through an importance ratio, assume an architecture whose exact likelihood is available, and are confined to de novo design. Whether such an update transfers across architectures whose training objectives are likelihood bounds rather than exact likelihoods, whether it extends to two-pass constrained generation procedures that require a dedicated joint-policy formulation, and whether weighting is needed once selection is in place, remain open questions.

%%%%%%%%%%%%%

\section{Implementation Details}
\label{app:impl}

\paragraph{Generators.} All three generative baselines used in this paper are pretrained. GenMol is the released \texttt{model\_v1} checkpoint \citep{lee2025genmol}. InVirtuoGen is the released \texttt{invirtuo\_gen} checkpoint \citep{kaech2026refine}. SAFE-GPT is loaded from HuggingFace \url{https://huggingface.co/datamol-io/safe-gpt}.

\paragraph{Native optimizers.}
SAFE-GPT's native optimizer is PPO with an adaptive KL penalty and no clipping. The paper
states this setting but does not release the implementation, so we reimplemented it from the paper
together with the adaptive-penalty variant of \citet{ppo}. We keep a batch size of $100$ and a learning rate of $1 \times 10^{-5}$ under AdamW, and take the rest from that variant. The penalty is $\beta \hat{D}_{\mathrm{KL}}(\pi_\theta \Vert \pi_{\mathrm{ref}})$ against the frozen
pretrained policy rather than against the previous iterate, estimated with the $k_3$
estimator and averaged per action token so that the target does not scale with molecule
length. The target is $0.03$ per token and $\beta$ starts at $0.1$. After each step $\beta$
doubles when the measured divergence exceeds the target by more than $1.5\times$ and halves
when it falls below the target divided by $1.5$, clamped to $[10^{-4},10]$. Advantages are
the batch-standardized oracle scores. Each step accumulates over micro-batches of $16$ and clips gradients at $1.0$.

We use GenMol's fragment-remasking GA from its PMO setup. We use the pretrained model under constraints to sample molecules and initialize a population of $100$ fragments, each scored by the mean score of the molecules that contain it. The length of each cold-start molecule is drawn from the ZINC250k length distribution \citep{zinc}. Each subsequent round samples $128$ parents and remasks one fragment region of each.
 
InVirtuoGen's is Genetic-PPO with the default configuration for PMO (learning rate
$1 \times 10^{-5}$, $50$ timesteps, $5$ reinforce steps per round, clipping $\epsilon=0.2$,
experience replay size $300$, population size $100$, genetic prompter and mutation on, no
prescreening).

\paragraph{Genetic operators.} Each model's population content decides what its
operator can select, and its generation interface decides how the selected material is
regenerated (Table~\ref{tab:ga-ops}). GenMol and SAFE-GPT sample $1{,}000$ molecules de novo and $500$ under fragment constraints as a cold start, both charged to the oracle budget, and keep a ranked set of $100$ cut fragments, each scored by the mean oracle score of the molecules it was cut
from, so their operators select fragments and the model has to assemble a molecule from
them. InVirtuoGen samples $110$ molecules de novo and $300$ under constraints, since the anchor filter discards most of the latter, and keeps the $100$ highest-scoring whole molecules. Its genetic prompter vocabulary retains the top $10$ of them, so its operator recombines complete molecules.

GenMol and SAFE-GPT both build a parent by attaching fragments and then have the model
regenerates part of it. In de novo design, both attach two population fragments into one
molecule. GenMol remasks one randomly chosen fragment region
of the parent with $5$ to $15$ mask tokens. SAFE-GPT re-cuts the joined molecule and
uses the largest open piece as the prompt to finish regeneration. Under fragment constraints, a single population seed is joined to the fixed anchor, so both operators become mutations of that seed. GenMol remasks a region as before and only keeps molecules that contain the anchor. SAFE-GPT re-cuts as in de novo but keeps the largest open piece that still contains the anchor.

InVirtuoGen recombines two population molecules at the fragment level, keeping all but the
last fragment of the first and the last fragment of the second, and the flow model samples
a completion conditioned on that token prompt. Under fragment constraints, the prompter is
switched off and the anchor alone forms the prompt, with its tokens written back at every
denoising step. InVirtuoGen also applies the atom-level graph-based genetic algorithm of
\citet{graphGA} to the highest-scoring molecules found so far, keeping mutants that retain
the anchor. That operator is the only chemical variation among the three models, the
others resampling their own generative distribution.

\begin{table}[ht]
\caption{\textbf{Genetic operator per model.} }
\label{tab:ga-ops}
\centering
\resizebox{\textwidth}{!}{
\begin{tabular}{lllll}
\toprule
& & \multicolumn{2}{c}{Attachment or recombination} & \\
\cmidrule(lr){3-4}
Model & Population & De novo & Fragment-constrained & Regeneration \\
\midrule
SAFE-GPT    & $100$ scored fragments & attach pair, re-cut
            & attach anchor $+$ one seed, re-cut keeping the anchor
            & complete the open prefix \\
GenMol      & $100$ scored fragments & attach fragment pair
            & attach anchor $+$ one seed
            & remask a random region, $5$--$15$ tokens \\
InVirtuoGen & $10$ scored molecules & fragment-level prompt crossover
            & no crossover; the anchor alone is the prompt
            & prompt-conditioned flow sampling, anchor forced \\
\bottomrule
\end{tabular}
}
\end{table}

\paragraph{EW-SFT.} For three models, the update is AdamW at learning rate $1 \times 10^{-5}$ with gradient clipping
at $1.0$. Each round samples $128$ molecules (GenMol, SAFE-GPT) or $100$ (InVirtuoGen, whose offspring count its own GA sets), scores
them, and takes one optimizer step in micro-batches of $16$ in rounds where the elite buffer changes. GenMol and SAFE-GPT use a rolling buffer of capacity $K\,{=}\,64$ to be the elite buffer, because their populations are fragment-level and retain no complete winner. InVirtuoGen realizes it through its own
genetic prompter vocabulary, which already holds whole molecules ranked by their oracle score, so no second buffer is added (Table~\ref{tab:ablation-design}).

\paragraph{Oracle.} The 3D shape similarity of ROSHAMBO2 \citep{roshambo2} is computed by Tanimoto Combination at its default equal weighting of shape and color, against a single reference conformer per reference compound. A molecule is scored by enumerating up to $4$ stereoisomers and $25$ ETKDGv3 conformers each and taking its best overlap with the reference. We fix the random seed to 42 to ensure reproducibility. Conformer generation dominates the computing cost, using multiple CPU cores to compute in parallel can accelerate it (Table~\ref{tab:runtime}). A molecule whose conformation embedding or alignment fails returns $-1$, it still consumes the oracle budget but the reported metrics drop it. Oracle calls are charged once per unique canonical SMILES. A repeat is queried from cache and costs no budget, so the budget counts distinct molecules evaluated. 

\paragraph{Early stopping and budget accounting.} InVirtuoGen's Genetic-PPO counts rounds in which fewer than $5\%$ of the generated molecules are new, discounting them by rounds in which at least half are, and stops when that count reaches $10$. Our reimplementation of SAFE-GPT PPO early stops if the collapsed policy continues generating cached duplicates over $30$ rounds. GenMol's GA and EW-SFT do not need such a mechanism, because GA draws from a population that keeps turning over, so its output keeps changing even as the policy narrows, whereas sampling straight from a narrowing policy re-emits the same molecules. A run that early stops is held flat at its final top-10 mean for the remainder, following the \emph{mol\_opt} convention, which neither rewards nor penalizes stopping. The Top-1k metric is computed over whatever the run produced.

\section{Additional Experimental Results}

\subsection{Fragment-constrained Anchors}
\label{app:anchors}

The four fragment-constrained-benchmark reference compounds are originally collected by \citet{safe-gpt}, and their anchors are obtained from \url{https://github.com/NVIDIA-BioNeMo/genmol/blob/main/data/fragments.csv}.

% ============================================================
% Fragment-constrained anchors (all six references)
% Depictions rendered by analysis/fragment_depictions.py
% ============================================================
\begin{table}[ht]
\caption{\textbf{Fragment-constrained anchors.} A motif-extension anchor has one
attachment point, the model grows the rest. Linker-design anchors contain two fixed end
fragments, the model designs the connecting core. Attachment points ($*$) are
highlighted in orange. Top panel: the two kinases used in the main experiments. Bottom panel:
the four fragment-constrained-benchmark references used for the held-out evaluation of
Section~\ref{sec:generalizability}.}
\label{tab:anchors}
\centering
\resizebox{0.8\textwidth}{!}{
\begin{tabular}{l ccc}
\toprule
Reference & Molecule & Motif anchor & Linker anchor \\
\midrule
BTK inhibitor &
  \raisebox{-.48\height}{\includegraphics[height=2.2cm]{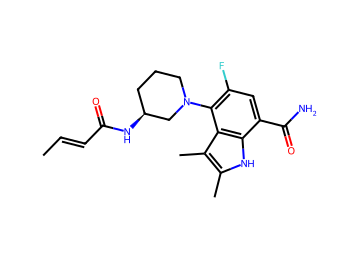}} &
  \raisebox{-.48\height}{\includegraphics[height=2.2cm]{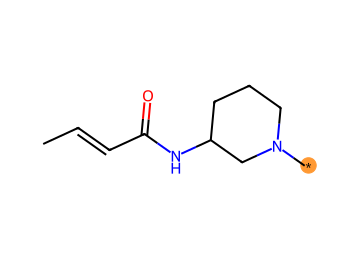}} &
  \raisebox{-.48\height}{\includegraphics[height=2.2cm]{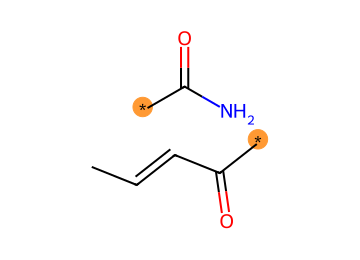}} \\[6pt]
TYK2 inhibitor &
  \raisebox{-.48\height}{\includegraphics[height=2.2cm]{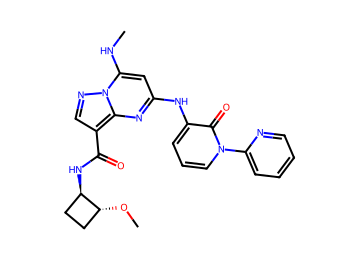}} &
  \raisebox{-.48\height}{\includegraphics[height=2.2cm]{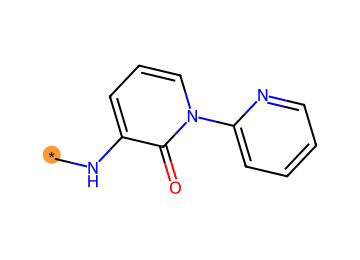}} &
  \raisebox{-.48\height}{\includegraphics[height=2.2cm]{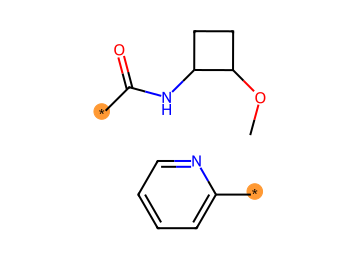}} \\
\midrule
Eliglustat &
  \raisebox{-.48\height}{\includegraphics[height=2.2cm]{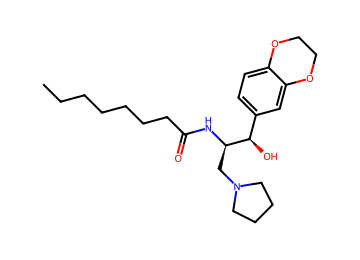}} &
  \raisebox{-.48\height}{\includegraphics[height=2.2cm]{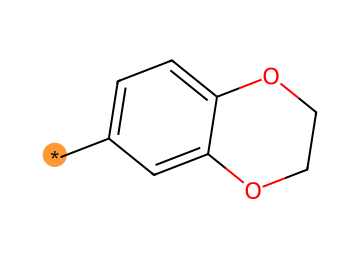}} &
  \raisebox{-.48\height}{\includegraphics[height=2.2cm]{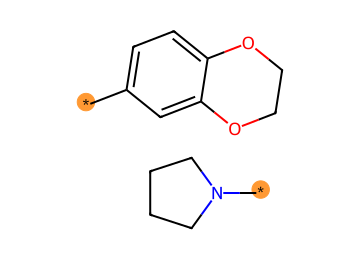}} \\[6pt]
Erlotinib &
  \raisebox{-.48\height}{\includegraphics[height=2.2cm]{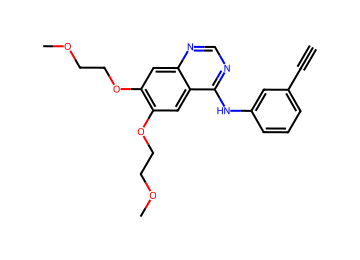}} &
  \raisebox{-.48\height}{\includegraphics[height=2.2cm]{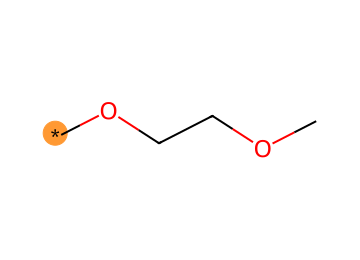}} &
  \raisebox{-.48\height}{\includegraphics[height=2.2cm]{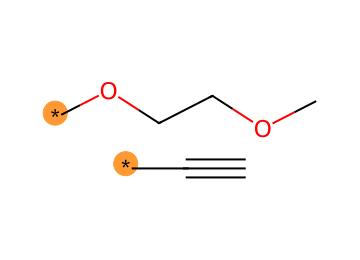}} \\[6pt]
Futibatinib &
  \raisebox{-.48\height}{\includegraphics[height=2.2cm]{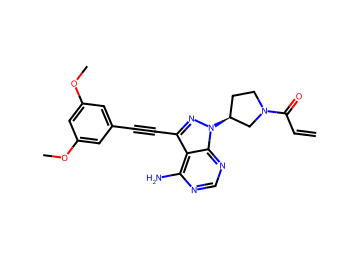}} &
  \raisebox{-.48\height}{\includegraphics[height=2.2cm]{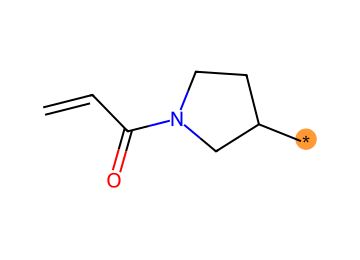}} &
  \raisebox{-.48\height}{\includegraphics[height=2.2cm]{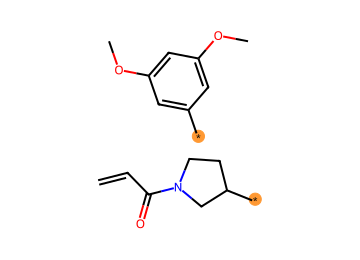}} \\[6pt]
Spirapril &
  \raisebox{-.48\height}{\includegraphics[height=2.2cm]{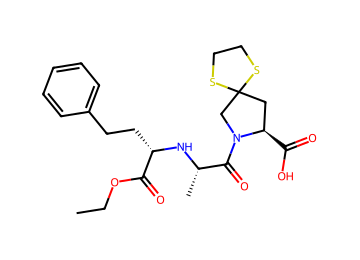}} &
  \raisebox{-.48\height}{\includegraphics[height=2.2cm]{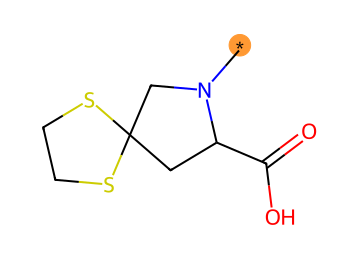}} &
  \raisebox{-.48\height}{\includegraphics[height=2.2cm]{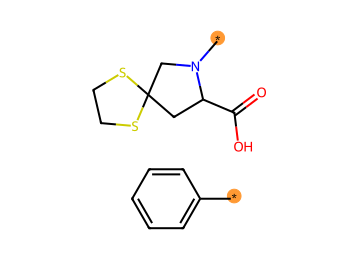}} \\
\bottomrule
\end{tabular}
}
\end{table}

\newpage
\subsection{Hypothesis Testing}
\label{app:stats}

Unless stated otherwise, each $p$-value we report is a Wilcoxon signed-rank test on paired differences, one-sided where the claim is directional (EW-SFT improves on the native optimizer, or the native optimizer improves on frozen sampling) and two-sided for the nondirectional PMO comparison. The reported $p$-values are Holm-adjusted within a family of hypotheses.

We pair hypothesis tests by seed, with the two arms sharing an initial population and other random states. Table~\ref{tab:main-top1k} pairs on (reference, task, seed), giving $n=18$ per model and $n=12$ for SAFE-GPT, because we do not evaluate PPO for its two-pass linker generation. Table~\ref{tab:generalizability} pairs the same way over four references and two tasks, $n=24$. The PMO comparison pairs on the oracle, $n=23$, because the competitor columns there are per-oracle means transcribed from the InVirtuoGen paper \citep{kaech2026refine}, and their individual runs are unavailable, so a seed-level pairing is not defined.

The quality--diversity claims in Section~\ref{sec:ablation} are the exception, since they concern a correlation rather than a paired comparison. For each of the $12$ cells (three models $\times$ two references $\times$ \{de novo, linker\}) we compute the Pearson correlation between Top-1k internal diversity and Top-1k score over the four optimizing arms and three seeds, $n=12$ points per cell, and test the resulting $12$ correlations against zero with a one-sample $t$-test. The SFT-only arm is excluded because it is a single extreme point (lowest score, highest diversity), and including it raises the average correlation to $-0.88$.

\subsection{Frozen Generation}
\label{app:frozen}

The frozen baseline of Section~\ref{sec:3d} is additionally characterized by the fraction of chemically valid and single connected molecules (validity), fraction of distinct canonical SMILES among the valid (uniqueness), and pairwise ECFP4 Tanimoto distance~\cite{rogers2010extended} over 1000 random molecules (diversity) \citep{safe-gpt}.

Table~\ref{tab:frozen-quality} shows all four frozen models produce near-perfectly valid and unique ($>0.926$) samples with high diversity ($>0.870$) for de novo design. Under fragment constraints, however, validity is largely preserved while uniqueness and diversity drop sharply. These constraints make generation more challenging by restricting the feasible design space to a narrow region of chemical space.
This degradation is both anchor- and model-specific. On linker design, both SAFE-GPT and InVirtuoGen generate fewer valid compounds on BTK task ($0.461$, $0.181$) than for TYK2 ($0.998$, $0.256$). InVirtuoGen fails differently due to its hard write-back for fragment anchor tokens \citep{kaech2026refine}, which may generate disconnected molecules, yielding only $0.083$ validity of BTK motif extension. These results show that frozen sampling remains reliable in the unconstrained setting but cannot transfer robustly to constrained design.

% ============================================================
% Table 1: Frozen sampling results (no optimization)
% ============================================================
\begin{table}[ht]
\caption{\textbf{Frozen sampling results.} The results are over 3 runs, higher is better.}
\label{tab:frozen-quality}
\centering
\resizebox{0.8\textwidth}{!}{
\begin{tabular}{llcccccc}
\toprule
& & \multicolumn{3}{c}{BTK} & \multicolumn{3}{c}{TYK2} \\
\cmidrule(lr){3-5} \cmidrule(lr){6-8}
Model & Task & Valid. & Uniq. & Div. & Valid. & Uniq. & Div. \\
\midrule
%% --- REINVENT ---
\multirow{3}{*}{REINVENT}
  & De novo  & 0.992{\scriptsize$\pm$.001} & 1.000{\scriptsize$\pm$.000} & 0.870{\scriptsize$\pm$.001} & 0.992{\scriptsize$\pm$.001} & 1.000{\scriptsize$\pm$.000} & 0.870{\scriptsize$\pm$.001} \\
  & Motif    & 1.000{\scriptsize$\pm$.000} & 0.345{\scriptsize$\pm$.001} & 0.569{\scriptsize$\pm$.001} & 1.000{\scriptsize$\pm$.000} & 0.697{\scriptsize$\pm$.005} & 0.587{\scriptsize$\pm$.001} \\
  & Linker   & 1.000{\scriptsize$\pm$.000} & 0.830{\scriptsize$\pm$.001} & 0.764{\scriptsize$\pm$.002} & 1.000{\scriptsize$\pm$.000} & 0.440{\scriptsize$\pm$.003} & 0.591{\scriptsize$\pm$.002} \\
\midrule
%% --- SAFE-GPT ---
\multirow{3}{*}{SAFE-GPT}
  & De novo  & 0.926{\scriptsize$\pm$.001} & 1.000{\scriptsize$\pm$.000} & 0.879{\scriptsize$\pm$.001} & 0.926{\scriptsize$\pm$.001} & 1.000{\scriptsize$\pm$.000} & 0.879{\scriptsize$\pm$.001} \\
  & Motif    & 0.967{\scriptsize$\pm$.001} & 0.206{\scriptsize$\pm$.001} & 0.503{\scriptsize$\pm$.006} & 0.946{\scriptsize$\pm$.001} & 0.253{\scriptsize$\pm$.005} & 0.517{\scriptsize$\pm$.001} \\
  & Linker   & 0.461{\scriptsize$\pm$.050} & 0.617{\scriptsize$\pm$.007} & 0.740{\scriptsize$\pm$.005} & 0.998{\scriptsize$\pm$.003} & 0.500{\scriptsize$\pm$.098} & 0.565{\scriptsize$\pm$.014} \\
\midrule
%% --- GenMol ---
\multirow{3}{*}{GenMol}
  & De novo  & 0.989{\scriptsize$\pm$.001} & 0.928{\scriptsize$\pm$.001} & 0.894{\scriptsize$\pm$.001} & 0.989{\scriptsize$\pm$.001} & 0.928{\scriptsize$\pm$.001} & 0.894{\scriptsize$\pm$.001} \\
  & Motif    & 0.989{\scriptsize$\pm$.001} & 0.241{\scriptsize$\pm$.003} & 0.558{\scriptsize$\pm$.001} & 0.985{\scriptsize$\pm$.001} & 0.195{\scriptsize$\pm$.001} & 0.533{\scriptsize$\pm$.002} \\
  & Linker   & 0.998{\scriptsize$\pm$.000} & 0.543{\scriptsize$\pm$.002} & 0.715{\scriptsize$\pm$.001} & 0.999{\scriptsize$\pm$.000} & 0.316{\scriptsize$\pm$.002} & 0.541{\scriptsize$\pm$.001} \\
\midrule
%% --- InVirtuoGen ---
\multirow{3}{*}{InVirtuoGen}
  & De novo  & 0.978{\scriptsize$\pm$.000} & 1.000{\scriptsize$\pm$.000} & 0.881{\scriptsize$\pm$.001} & 0.978{\scriptsize$\pm$.000} & 1.000{\scriptsize$\pm$.000} & 0.881{\scriptsize$\pm$.001} \\
  & Motif    & 0.083{\scriptsize$\pm$.004} & 0.852{\scriptsize$\pm$.010} & 0.566{\scriptsize$\pm$.001} & 0.853{\scriptsize$\pm$.002} & 0.578{\scriptsize$\pm$.005} & 0.535{\scriptsize$\pm$.000} \\
  & Linker   & 0.181{\scriptsize$\pm$.002} & 0.872{\scriptsize$\pm$.003} & 0.762{\scriptsize$\pm$.001} & 0.256{\scriptsize$\pm$.003} & 0.693{\scriptsize$\pm$.008} & 0.548{\scriptsize$\pm$.001} \\
\bottomrule
\end{tabular}
}
\end{table}

\subsection{Gain Decomposition in Absolute Units}
\label{app:gain}

Figure~\ref{fig:gain} reports each term as a fraction of the cell's frozen baseline. Table~\ref{tab:gain} shows the decomposition in raw Top-1k absolute gain.

% Table 3: Gain decomposition
% ============================================================
\begin{table}[ht]
\caption{\textbf{Optimization improvement over different settings.} The results are Top-1k mean 3D similarity over 3 runs. $\overline{\Delta}_{\mathrm{nat}-\mathrm{frz}}$ indicates what the model's native optimizer adds over frozen sampling, $\overline{\Delta}_{\mathrm{ours}-\mathrm{nat}}$ is the EW-SFT-to-native improvement, and $\overline{\Delta}_{\mathrm{ours}-\mathrm{frz}}$ shows what EW-SFT adds on the frozen sampling. SAFE-GPT linker has no native PPO arm because we do not implement the joint policy required for its two-pass generation.}
\label{tab:gain}
\centering
\resizebox{\textwidth}{!}{
\begin{tabular}{ll cccccc |c}
\toprule
& & \multicolumn{3}{c}{BTK} & \multicolumn{3}{c}{TYK2} & \\
\cmidrule(lr){3-5} \cmidrule(lr){6-8}
Model & Term & De novo & Motif & Linker & De novo & Motif & Linker & Avg. \\
\midrule
\multirow{4}{*}{SAFE-GPT}
  & Frozen (Top-1k) & 0.521 & 0.534 & 0.502 & 0.400 & 0.428 & 0.406 & 0.465 \\
  & $\overline{\Delta}_{\mathrm{nat}-\mathrm{frz}}$ & $+$0.007 & $+$0.038 & --- & $+$0.008 & $+$0.031 & --- & $+$0.021 \\
  & $\overline{\Delta}_{\mathrm{ours}-\mathrm{nat}}$ & $+$0.102 & $+$0.093 & --- & $+$0.136 & $+$0.033 & --- & $+$0.091 \\
  & \oc $\overline{\Delta}_{\mathrm{ours}-\mathrm{frz}}$ & \oc $+$0.109 & \oc $+$0.131 & \oc $+$0.061 & \oc $+$0.144 & \oc $+$0.064 & \oc $+$0.039 & \oc $+$0.091 \\
\midrule
\multirow{4}{*}{GenMol}
  & Frozen (Top-1k) & 0.525 & 0.594 & 0.560 & 0.381 & 0.415 & 0.386 & 0.477 \\
  & $\overline{\Delta}_{\mathrm{nat}-\mathrm{frz}}$ & $+$0.096 & $+$0.063 & $+$0.064 & $+$0.112 & $+$0.071 & $+$0.135 & $+$0.090 \\
  & $\overline{\Delta}_{\mathrm{ours}-\mathrm{nat}}$ & $+$0.009 & $+$0.006 & $+$0.026 & $+$0.018 & $+$0.003 & $+$0.011 & $+$0.012 \\
  & \oc $\overline{\Delta}_{\mathrm{ours}-\mathrm{frz}}$ & \oc $+$0.105 & \oc $+$0.068 & \oc $+$0.090 & \oc $+$0.130 & \oc $+$0.075 & \oc $+$0.146 & \oc $+$0.102 \\
\midrule
\multirow{4}{*}{InVirtuoGen}
  & Frozen (Top-1k) & 0.531 & 0.564 & 0.518 & 0.404 & 0.440 & 0.376 & 0.472 \\
  & $\overline{\Delta}_{\mathrm{nat}-\mathrm{frz}}$ & $+$0.192 & $+$0.088 & $+$0.139 & $+$0.161 & $+$0.068 & $+$0.130 & $+$0.129 \\
  & $\overline{\Delta}_{\mathrm{ours}-\mathrm{nat}}$ & $+$0.074 & $+$0.117 & $+$0.113 & $+$0.047 & $+$0.052 & $+$0.072 & $+$0.079 \\
  & \oc $\overline{\Delta}_{\mathrm{ours}-\mathrm{frz}}$ & \oc $+$0.266 & \oc $+$0.205 & \oc $+$0.251 & \oc $+$0.208 & \oc $+$0.119 & \oc $+$0.202 & \oc $+$0.209 \\
\bottomrule
\end{tabular}
}
\end{table}

\newpage
\subsection{Optimization Convergence Curves and AUC}
\label{app:convergence}

\begin{figure}[ht]
\centering
\includegraphics[width=\textwidth]{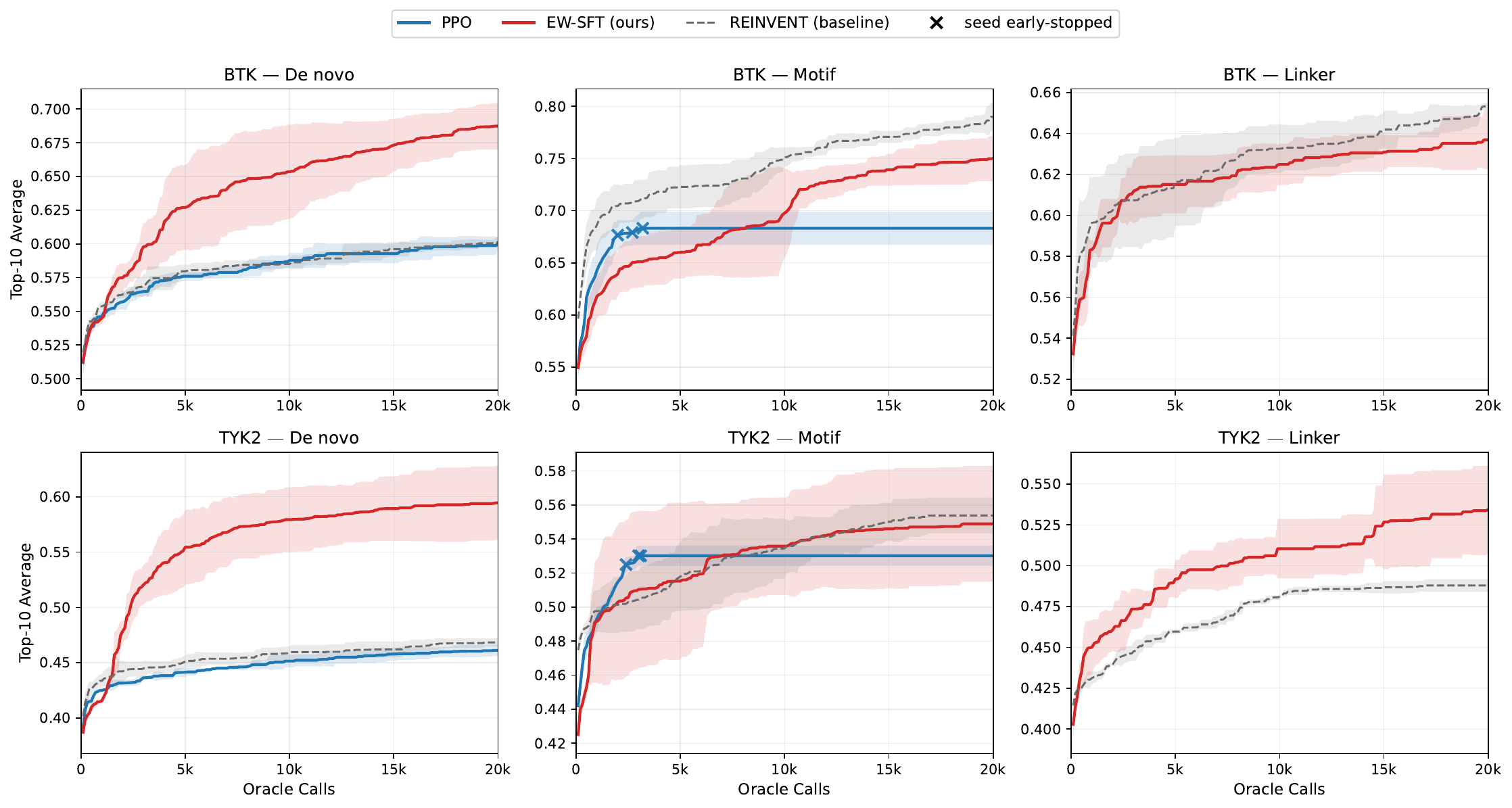}
\caption{\textbf{Optimization convergence for SAFE-GPT.} SAFE-GPT's native PPO with adaptive KL penalty vs.\ EW-SFT (ours), with REINVENT overlaid in grey as an external baseline. An $\times$ marks where a run stopped, the curves are then held flat, so the blue curve is truncated rather than plateauing. EW-SFT runs the full budget in every panel. The linker column has no PPO curve as discussed in Section~\ref{sec:prelim}.}
\label{fig:safe-convergence}
\end{figure}

\begin{figure}[ht]
\centering
\includegraphics[width=\textwidth]{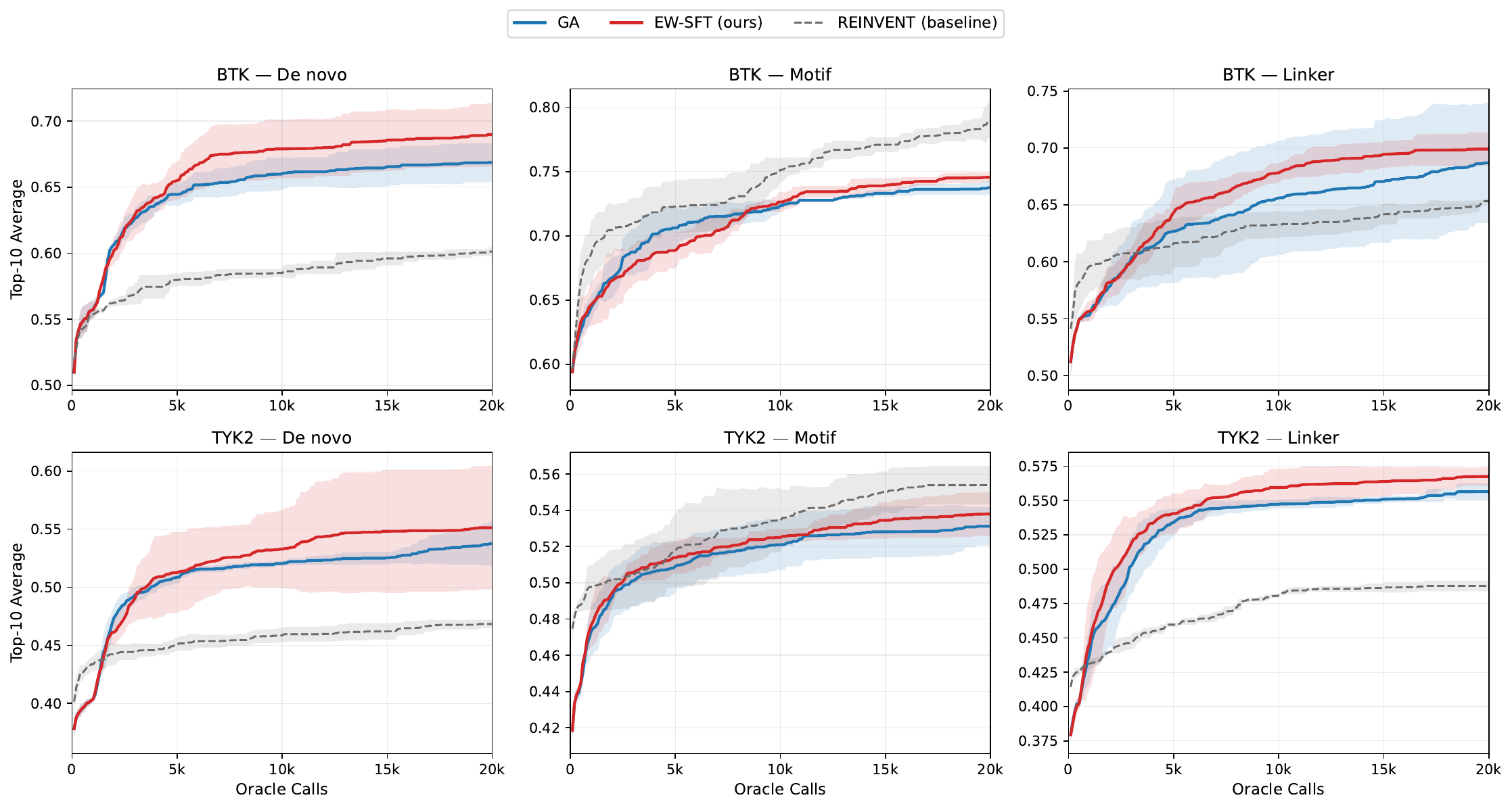}
\caption{\textbf{Optimization convergence for GenMol.} GenMol's native fragment-remasking GA vs.\ EW-SFT (ours), with REINVENT overlaid in grey as an external baseline. Neither GenMol optimizer early-stops, both curves run to the full budget.}
\label{fig:genmol-convergence}
\end{figure}

\begin{figure}[ht]
\centering
\includegraphics[width=\textwidth]{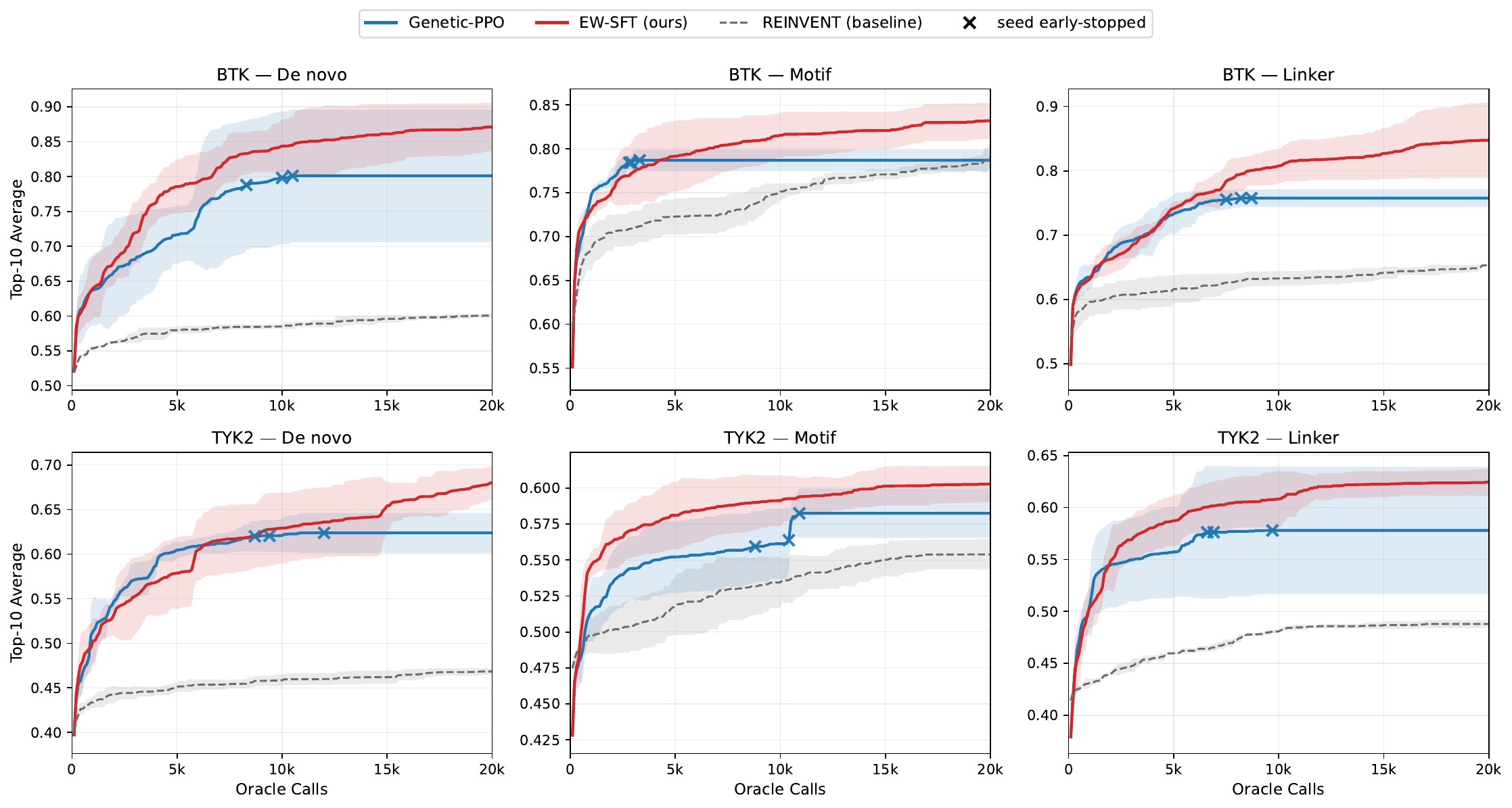}
\caption{\textbf{Optimization convergence for InVirtuoGen.} InVirtuoGen's native Genetic-PPO vs.\ EW-SFT
(ours), with REINVENT overlaid in grey as an external baseline. An $\times$ marks where a run early-stopped. Genetic-PPO exhausts its novelty budget and stops early, whereas EW-SFT keeps producing novel candidates through the full budget.}
\label{fig:ivg-convergence}
\end{figure}

Figures~\ref{fig:safe-convergence}-~\ref{fig:ivg-convergence} illustrate the running Top-10 mean versus oracle calls for each model's native optimizer and its EW-SFT replacement. Table~\ref{tab:app-auc} shows the corresponding AUC Top-10 results.

% ============================================================
% Appendix Table: AUC Top-10 (mean ± std, 3 seeds)
% ============================================================
\begin{table}[ht]
\caption{\textbf{AUC Top-10 scores of 3D shape similarity optimization.} The results are mean$\pm$std over 3 runs, with 20k oracle calls. SAFE-GPT and InVirtuoGen arms inherit their models' native early stopping. Following the \emph{mol\_opt} convention a run that ends early contributes its final Top-10 mean over the remaining calls, which neither rewards nor penalizes stopping early. The final area under the running Top-10 curve is normalized by the full budget.}
\label{tab:app-auc}
\centering
\resizebox{\textwidth}{!}{
\begin{tabular}{ll cccccc |c}
\toprule
& & \multicolumn{3}{c}{BTK} & \multicolumn{3}{c}{TYK2} & \\
\cmidrule(lr){3-5} \cmidrule(lr){6-8}
Model & Method & De novo & Motif & Linker & De novo & Motif & Linker & Avg. \\
\midrule
\multirow{1}{*}{REINVENT}
  & REINFORCE & 0.583{\scriptsize$\pm$0.004} & 0.741{\scriptsize$\pm$0.011} & 0.626{\scriptsize$\pm$0.013} & 0.454{\scriptsize$\pm$0.006} & 0.530{\scriptsize$\pm$0.013} & 0.470{\scriptsize$\pm$0.001} & 0.567 \\
\midrule
\multirow{2}{*}{SAFE-GPT}
  & PPO & 0.581{\scriptsize$\pm$0.005} & 0.676{\scriptsize$\pm$0.016} & --- & 0.447{\scriptsize$\pm$0.004} & 0.525{\scriptsize$\pm$0.005} & --- & 0.557 \\
  & \oc EW-SFT (ours) & \oc \best{0.641}{\scriptsize$\pm$0.022} & \oc \best{0.695}{\scriptsize$\pm$0.017} & \oc \best{0.618}{\scriptsize$\pm$0.010} & \oc \best{0.557}{\scriptsize$\pm$0.026} & \oc \best{0.528}{\scriptsize$\pm$0.034} & \oc \best{0.501}{\scriptsize$\pm$0.015} & \oc \best{0.590} \\
\midrule
\multirow{2}{*}{GenMol}
  & GA & 0.645{\scriptsize$\pm$0.011} & 0.711{\scriptsize$\pm$0.005} & 0.641{\scriptsize$\pm$0.044} & 0.508{\scriptsize$\pm$0.002} & 0.514{\scriptsize$\pm$0.013} & 0.530{\scriptsize$\pm$0.007} & 0.592 \\
  & \oc EW-SFT (ours) & \oc \best{0.660}{\scriptsize$\pm$0.018} & \oc \best{0.711}{\scriptsize$\pm$0.005} & \oc \best{0.658}{\scriptsize$\pm$0.009} & \oc \best{0.519}{\scriptsize$\pm$0.037} & \oc \best{0.518}{\scriptsize$\pm$0.006} & \oc \best{0.542}{\scriptsize$\pm$0.011} & \oc \best{0.602} \\
\midrule
\multirow{2}{*}{InVirtuoGen}
  & Genetic-PPO & 0.758{\scriptsize$\pm$0.080} & 0.778{\scriptsize$\pm$0.011} & 0.734{\scriptsize$\pm$0.015} & 0.601{\scriptsize$\pm$0.012} & 0.561{\scriptsize$\pm$0.010} & 0.564{\scriptsize$\pm$0.054} & 0.666 \\
  & \oc EW-SFT (ours) & \oc \best{0.807}{\scriptsize$\pm$0.028} & \oc \best{0.799}{\scriptsize$\pm$0.024} & \oc \best{0.776}{\scriptsize$\pm$0.026} & \oc \best{0.612}{\scriptsize$\pm$0.025} & \oc \best{0.584}{\scriptsize$\pm$0.016} & \oc \best{0.595}{\scriptsize$\pm$0.013} & \oc \best{0.695} \\
\bottomrule
\end{tabular}
}
\end{table}

\newpage
\subsection{2D Novelty}
\label{app:scaffold}

On the same Top-1k molecules scored by 3D shape similarity against each of the two kinase references in Table~\ref{tab:main-top1k}, we also compute the ECFP4 Tanimoto similarity of each molecule to the reference (Table~\ref{tab:novelty2d}), to show whether the optimizer improves the 3D score by rediscovering the reference compound.

% ============================================================
% Appendix table: 2D similarity of the reported molecules
% ============================================================
\begin{table}[ht]
\caption{\textbf{2D similarity to the reference.} The results are ECFP4 Tanimoto similarity over the Top-1k by 3D score (mean $\pm$ std over 3 runs). A molecule that has high 3D shape similarity while low 2D similarity to reference means it is shape-matched but chemically distinct. The motif and linker columns have higher scores because the provided anchor is a fragment of the reference and already supplies some similarity. \underline{Underlined} EW-SFT cell is less similar to the reference than the model's native optimizer ($9$ of $16$ cells). Lower is the desirable direction here.}
\label{tab:novelty2d}
\centering
\resizebox{\textwidth}{!}{
\begin{tabular}{llcccccc}
\toprule
& & \multicolumn{3}{c}{BTK} & \multicolumn{3}{c}{TYK2} \\
\cmidrule(lr){3-5} \cmidrule(lr){6-8}
Model & Method & De novo & Motif & Linker & De novo & Motif & Linker \\
\midrule
\multirow{2}{*}{SAFE-GPT}
  & PPO & 0.125{\scriptsize$\pm$0.001} & 0.394{\scriptsize$\pm$0.009} & --- & 0.139{\scriptsize$\pm$0.002} & 0.361{\scriptsize$\pm$0.002} & --- \\
  & \oc EW-SFT (ours) & \oc 0.159{\scriptsize$\pm$0.085} & \oc 0.438{\scriptsize$\pm$0.018} & \oc 0.207{\scriptsize$\pm$0.005} & \oc 0.153{\scriptsize$\pm$0.017} & \oc \underline{0.310}{\scriptsize$\pm$0.011} & \oc 0.328{\scriptsize$\pm$0.005} \\
\midrule
\multirow{2}{*}{GenMol}
  & GA & 0.133{\scriptsize$\pm$0.020} & 0.411{\scriptsize$\pm$0.013} & 0.229{\scriptsize$\pm$0.061} & 0.138{\scriptsize$\pm$0.014} & 0.347{\scriptsize$\pm$0.012} & 0.350{\scriptsize$\pm$0.007} \\
  & \oc EW-SFT (ours) & \oc 0.157{\scriptsize$\pm$0.062} & \oc \underline{0.405}{\scriptsize$\pm$0.006} & \oc \underline{0.190}{\scriptsize$\pm$0.018} & \oc 0.157{\scriptsize$\pm$0.025} & \oc \underline{0.347}{\scriptsize$\pm$0.023} & \oc \underline{0.341}{\scriptsize$\pm$0.006} \\
\midrule
\multirow{2}{*}{InVirtuoGen}
  & Genetic-PPO & 0.170{\scriptsize$\pm$0.097} & 0.372{\scriptsize$\pm$0.016} & 0.163{\scriptsize$\pm$0.028} & 0.148{\scriptsize$\pm$0.032} & 0.238{\scriptsize$\pm$0.029} & 0.193{\scriptsize$\pm$0.044} \\
  & \oc EW-SFT (ours) & \oc 0.249{\scriptsize$\pm$0.073} & \oc \underline{0.279}{\scriptsize$\pm$0.075} & \oc 0.254{\scriptsize$\pm$0.087} & \oc \underline{0.103}{\scriptsize$\pm$0.029} & \oc \underline{0.229}{\scriptsize$\pm$0.063} & \oc \underline{0.186}{\scriptsize$\pm$0.052} \\
\bottomrule
\end{tabular}
}
\end{table}

\subsection{Ablation Settings}
\label{app:ablation-design}

Table~\ref{tab:ablation-design} defines the six ablation settings compared in Section~\ref{sec:ablation} Table~\ref{tab:ablation-results}. As genetic search+elite+binary is the best setting for EW-SFT, as a comparison, the SFT-only setting is also based on elite+binary but removing the genetic search and sampling straight from the pretrained model. Proposition~\ref{prop:unbounded} shows the signed emphasis is ill-posed and it empirically does not perform well, so we exclude the elite+signed setting. The batch-binary is also excluded because the reward signal needs to enter somewhere. The elite buffer differs by architecture, but every model is run at exactly one molecule-level elite stage. GenMol and SAFE-GPT evolve fragment populations with no complete winners, so for them the buffer has to be added. In contrast, InVirtuoGen runs on the round batch, because its genetic prompter vocabulary already keeps whole molecules ranked by their own oracle score, which is what $E_t$ is defined to be. Its $\tau_t\,{=}\,-\infty$ rows are therefore its elite rows, and adding our buffer on top would compose a second selection stage.

% ============================================================
% Table: Ablation design
% ============================================================
\begin{table}[ht]
\caption{\textbf{Ablation settings.} Two factors of Equation~\ref{eq:weight-factor}: the threshold $\tau_t$ determines the selected set, and the emphasis re-weights the selected molecules. Elite is the rolling buffer of capacity $K\,{=}\,64$ with $\tau_t=\min_{x' \in E_t} f(x')$. $\tau_t=-\infty$ selects the whole valid round batch. GA is the genetic search of Section~\ref{sec:method}.}
\label{tab:ablation-design}
\centering
\begin{tabular}{lccc}
\toprule
Setting & GA & $\tau_t$ (threshold) & Emphasis \\
\midrule
\multicolumn{4}{l}{EW-SFT (ours)} \\
\quad elite + binary     & \checkmark & elite min   & $1$ \\
\quad elite + positive   & \checkmark & elite min    & $\max(\hat{A},0)$ \\
\quad batch + positive     & \checkmark & $-\infty$ & $\max(\hat{A},0)$ \\
\quad batch + signed & \checkmark & $-\infty$ & $\hat{A}$ \\
\midrule
\multicolumn{4}{l}{Standalone} \\
\quad GA only                & \checkmark & --- & --- \\
\quad SFT only                & ---        & elite min   & $1$ \\
\bottomrule
\end{tabular}
\end{table}

\subsection{Why the emphasis is not signed.}
\label{app:signed-emph}

\begin{figure}[ht]
    \centering
    \includegraphics[width=0.5\textwidth]{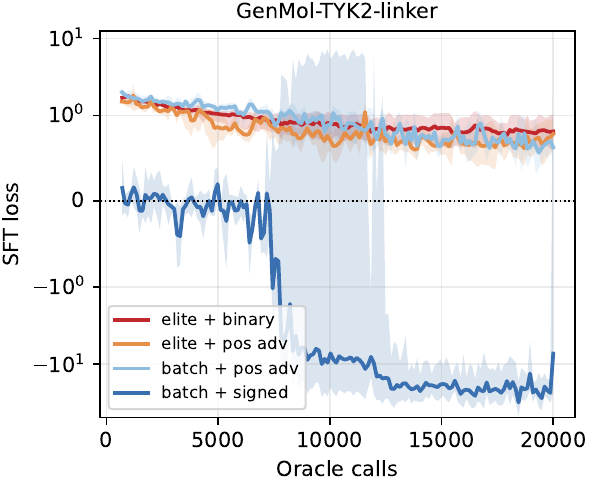}
    \caption{\textbf{Signed emphasis drives the SFT loss below zero.} The signed per-round training loss crosses zero and keeps descending, every other setting stays bounded below.}
    \label{fig:signed-divergence}
\end{figure}

A variant sets the emphasis to a batch-normalized advantage $\hat{A}$, which is negative for below-average samples. It is not merely worse empirically, but also unbounded. Gradient descent can therefore reduce the unbounded-below loss by pushing away disfavoured samples rather than toward the good ones. Figure~\ref{fig:signed-divergence} shows the failure mode. This update rule's property applies to all three models.

\begin{proposition}[Signed emphasis makes the objective unbounded below]
\label{prop:unbounded}
Fix a batch $\tilde{B}_t$ under signed emphasis, and suppose that
$w(x^-)=\hat{A}(x^-)<0$ for some $x^-\in\tilde{B}_t$. If the native loss
$\ell_\theta(x^-)$ is non-negative and unbounded above, then the weighted
objective
$
\sum_{x \in \tilde{B}_t} w(x)\,\ell_\theta(x)
$
is unbounded below. Holding the remaining terms finite, letting
$\ell_\theta(x^-)\to\infty$ gives
$w(x^-)\ell_\theta(x^-)\to-\infty$. Restricting to $w\geq0$ makes every term a
non-negative multiple of a non-negative loss, so the objective is bounded below
by $0$.
\end{proposition}

\subsection{Exploration and Exploitation Under the Elite Update}
\label{app:stall}

\begin{figure}[ht]
\centering
\includegraphics[width=\textwidth]{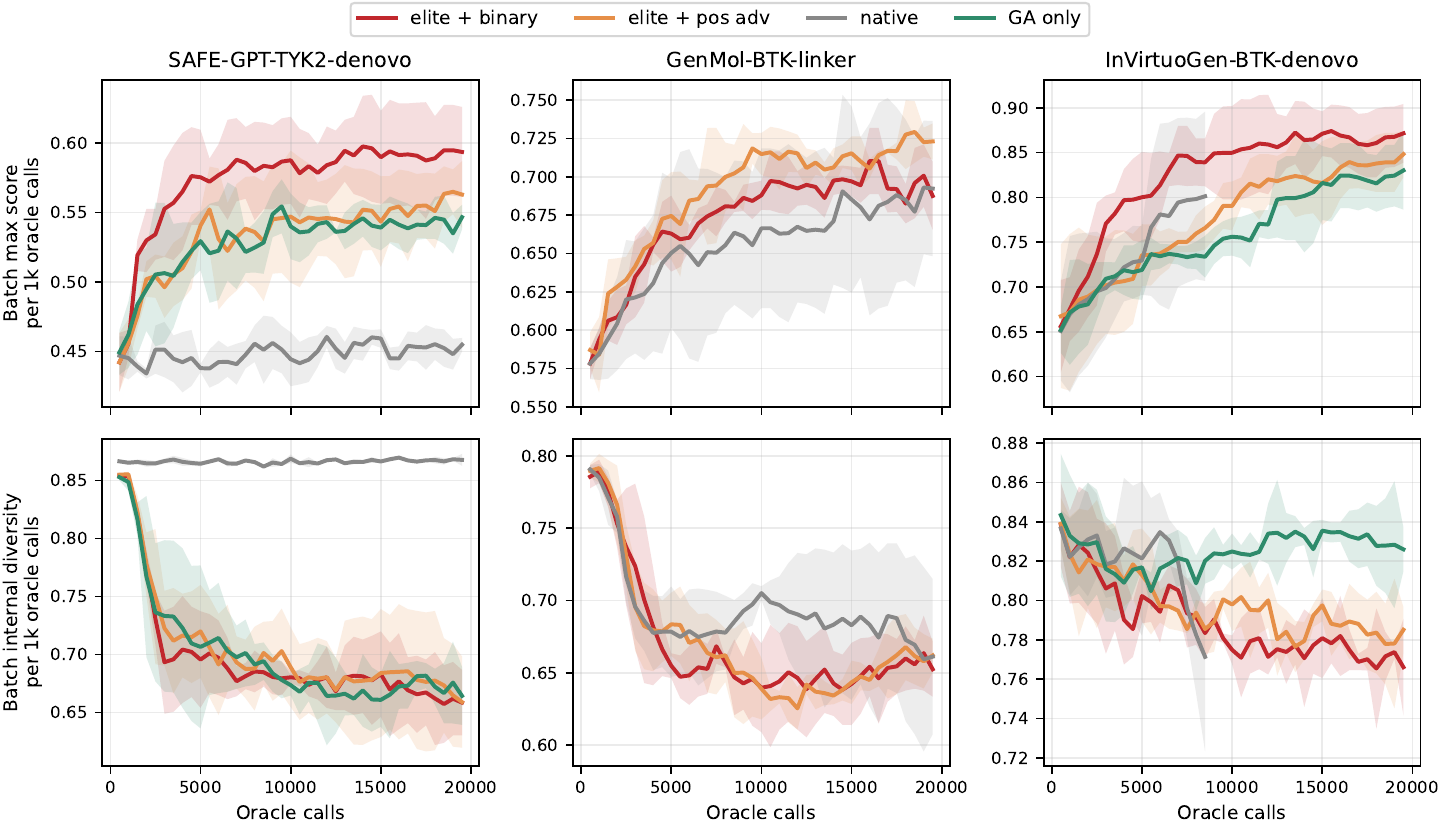}
\caption{\textbf{The exploration and exploitation dynamics of three optimization cases.} Each curve's point is computed within a rolling window of $1000$ oracle calls at a stride of $500$, over three seeds. Internal diversity is the pairwise ECFP4 Tanimoto distance within a window. InVirtuoGen's native Genetic-PPO curve early-stopped at ${\approx}8$k calls. GenMol's native optimizer is GA itself, so its grey curve is also its GA-only baseline.}
\label{fig:exploit-explore}
\end{figure}

Figure~\ref{fig:exploit-explore} shows the exploration and exploitation dynamics of three representative cases. EW-SFT's gain is associated with a longer productive search and is not accompanied by a significant reduction in diversity.

\subsection{Runtime}
\label{app:runtime}

Wall-clock hours reflect the budget each arm actually consumed, since the native optimizers frequently terminate before exhausting the
budget (Appendix~\ref{app:impl}).

% ============================================================
% Appendix: wall-clock cost
% ============================================================
\begin{table}[ht]
\caption{\textbf{Cost per run.} Wall-clock hours (h), thousands of oracle calls spent (K),
and their ratio (ms/call), as means over both references and three seeds on one A100 with 15
CPU. SAFE-GPT PPO has no linker entry because it is not evaluated on that task (Table~\ref{tab:main-top1k}).}
\label{tab:runtime}
\centering
\resizebox{\textwidth}{!}{
\begin{tabular}{ll ccc ccc ccc}
\toprule
& & \multicolumn{3}{c}{De novo} & \multicolumn{3}{c}{Motif} & \multicolumn{3}{c}{Linker} \\
\cmidrule(lr){3-5} \cmidrule(lr){6-8} \cmidrule(lr){9-11}
Model & Method & h & calls (K) & ms/call & h & calls (K) & ms/call & h & calls (K) & ms/call \\
\midrule
\multirow{2}{*}{SAFE-GPT}
  & Native & 1.5 & 20.0 & 278 & 1.1 & 2.7 & 1452 & --- & --- & --- \\
  & \oc EW-SFT (ours) & \oc 3.8 & \oc 20.0 & \oc 682 & \oc 4.9 & \oc 20.0 & \oc 883 & \oc 5.9 & \oc 20.0 & \oc 1056 \\
\midrule
\multirow{2}{*}{GenMol}
  & Native & 4.1 & 20.0 & 743 & 7.8 & 20.0 & 1411 & 8.8 & 20.0 & 1578 \\
  & \oc EW-SFT (ours) & \oc 5.4 & \oc 20.0 & \oc 964 & \oc 7.7 & \oc 20.0 & \oc 1394 & \oc 15.9 & \oc 20.0 & \oc 2868 \\
\midrule
\multirow{2}{*}{InVirtuoGen}
  & Native & 7.1 & 9.8 & 2618 & 3.0 & 6.5 & 1658 & 4.7 & 7.9 & 2151 \\
  & \oc EW-SFT (ours) & \oc 7.4 & \oc 20.0 & \oc 1325 & \oc 7.4 & \oc 20.0 & \oc 1327 & \oc 9.5 & \oc 20.0 & \oc 1703 \\
\bottomrule
\end{tabular}
}
\end{table}

\end{document}